\documentclass{article}

\usepackage[final]{neurips_2025}

\usepackage[utf8]{inputenc} 
\usepackage{graphicx}
\usepackage{booktabs}
\usepackage{epsfig} 
\usepackage{makecell}
\usepackage{times}
\usepackage{silence}
\usepackage{epstopdf}
\usepackage{color, colortbl}
\usepackage{float}  
\usepackage{caption}
\usepackage{multirow} 
\usepackage{multicol}
\usepackage{xcolor}
\usepackage{soul}
\usepackage{url}
\usepackage{placeins}
\usepackage[group-separator={,}]{siunitx}
\usepackage[utf8]{inputenc}
\usepackage{amssymb}
\usepackage{pifont}
\usepackage{paralist}
\usepackage{xspace}
\usepackage{tcolorbox}

\sethlcolor{orange!20} 
\newcommand{\hlgreen}[1]{\sethlcolor{green!20}\hl{#1}\sethlcolor{green!20}}
\newcommand{\hlred}[1]{\sethlcolor{cyan!20}\hl{#1}\sethlcolor{cyan!20}}

\newcommand{\vspaceundertab}{\vspace{0.1cm}}

\definecolor{LightCyan}{rgb}{0.88,1,1}
\definecolor{Gray}{gray}{0.9}
\definecolor{Pink}{rgb}{1, 0, 1}
\definecolor{azure}{rgb}{0.0, 0.44, 1.0}
\definecolor{bleudefrance}{rgb}{0.19, 0.55, 0.91}
\definecolor{cobalt}{rgb}{0.0, 0.28, 0.67}
\definecolor{electricpurple}{rgb}{0.75, 0.0, 1.0}

\newcommand{\re}[1]{\textcolor{red}{#1}}

\newcommand{\bl}[1]{\textcolor{blue}{#1}}

\newcommand{\epp}[1]{\textcolor{electricpurple}{#1}}

\definecolor{LightCyan}{rgb}{0.88,1,1}

\newcommand{\modelname}{VisPer-LM\xspace}
\newcommand\changes[1]{{#1}}

\definecolor{cvprblue}{rgb}{0.21,0.49,0.74}
\usepackage[pagebackref,breaklinks,colorlinks,
            linkcolor=cvprblue,
            citecolor=cvprblue,
            urlcolor=magenta]{hyperref}

\usepackage[capitalize]{cleveref}
\crefname{section}{Sec.}{Secs.}
\Crefname{section}{Section}{Sections}
\Crefname{table}{Table}{Tables}
\crefname{table}{Tab.}{Tabs.}

\title{Elevating Visual Perception in Multimodal LLMs with Visual Embedding Distillation}

\author{
  Jitesh Jain\textsuperscript{1,2}\thanks{Work done during JJ, JY's time with Microsoft Research.\textsuperscript{$\dagger$}Equal advising.} \quad
  Zhengyuan Yang\textsuperscript{2} \quad
  Humphrey Shi\textsuperscript{1}\textsuperscript{$\dagger$} \quad
  Jianfeng Gao\textsuperscript{2}\textsuperscript{$\dagger$} \quad
  Jianwei Yang\textsuperscript{3}\textsuperscript{$*\dagger$} \\[0.3em]
  \textsuperscript{1}SHI Labs @ Georgia Tech \quad 
  \textsuperscript{2}Microsoft Research, Redmond \quad \textsuperscript{3}Meta Superintelligence Labs \\[0.3em]
  \textbf{\texttt{\url{https://github.com/SHI-Labs/VisPer-LM}}}
}

\begin{document}

\maketitle

\begin{center}
    \centering
    \captionsetup{type=figure}
    \vspace{-0.6cm}
    \includegraphics[width=\textwidth]{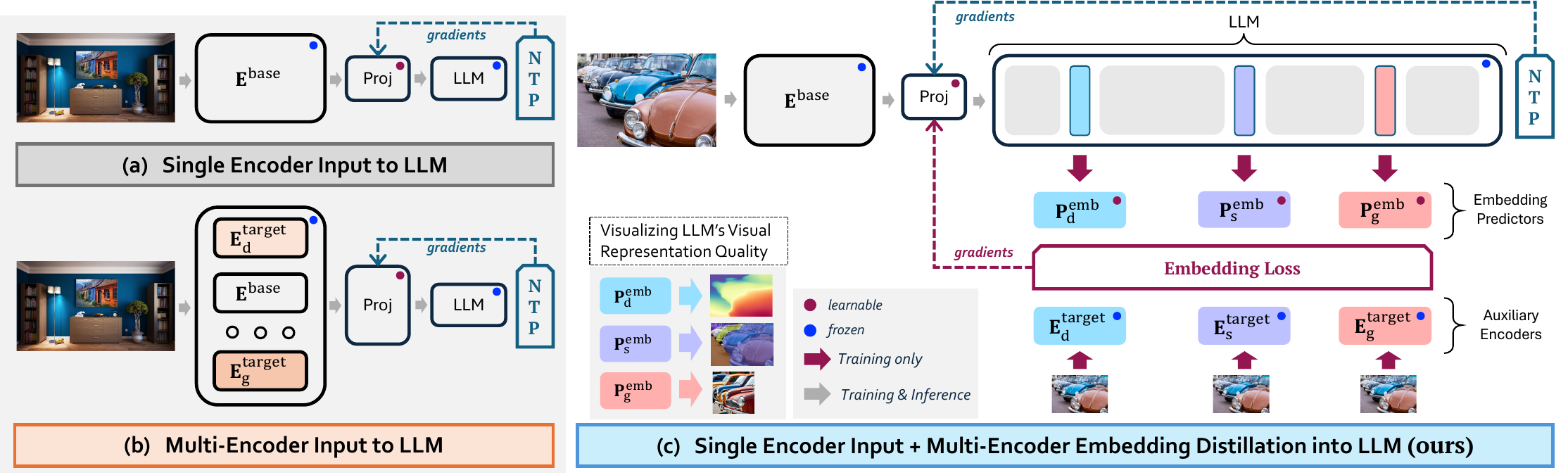}
    \vspace{-0.5cm}
    \captionof{figure}{\textbf{Different Paradigms for Incorporating Visual Information into LLMs.} \textbf{(a, b)} Existing approaches~\cite{liu2023improvedllava, tong2024cambrian1} feed features from the visual encoder(s) into the LLM and train the model solely with natural language supervision, i.e., next token prediction (NTP) to align the embedding space of the vision encoder(s) and the LLM. \textbf{(c)} We propose distilling target visual information into the intermediate representations of the LLM from a set of auxiliary vision encoders ($\mathbf{E}^\text{target}$). We adopt a predictive embedding~\cite{jepa} optimization approach at selected LLM layers during training to minimize the embedding losses and the NTP loss function, resulting in a vision-centric approach to training the Multimodal Large Language Model. We only use a single base vision encoder during inference.}
    \label{fig:teaser}
\end{center}%

\begin{abstract}
\vspace{-0.3cm}

\noindent
In recent times, the standard practice for developing MLLMs is to feed features from vision encoder(s) into the LLM and train with natural language supervision. This approach often causes models to lean towards language comprehension and undermine the rich visual perception signals present in the data, which are critical for tasks involving spatial reasoning in the domain of embodied AI and robotics. Is it possible to optimize both at the same time?
In this work, we propose \textbf{\modelname}, the first approach that infuses visual perception knowledge from expert vision encoders into the LLM's (of an MLLM) hidden representations. We start by investigating MLLMs trained solely with natural language supervision and identify a positive correlation between the quality of visual representations within these models and their downstream performance. Given this insight, we formulate the objective during the pretraining stage in MLLMs as a coupled optimization of predictive visual embedding and next (text) token prediction.  Moreover, through extensive probing, we observe improved visual representation quality due to embedding optimization, underscoring the effectiveness of our probing setup. We demonstrate that our \modelname outperforms the single and multi-encoder baselines, proving our approach's superiority over explicitly feeding the corresponding features to the LLM. In particular, \modelname boosts performance by an average margin of up to \textbf{2.5}\% on various benchmarks, with a notable improvement of \textbf{8.7}\% on the Depth task in CV-Bench.

\end{abstract}
\section{Introduction}
\vspace{-0.2cm}

\begin{figure}[t!]
\centering
\begin{minipage}{0.45\linewidth}
    \caption{\textbf{Probing reveals a positive correlation between depth representation quality and performance on CV-Bench.} 
    \textbf{(a)} Increasing training data and using only the next-token prediction objective improves visual representation quality in the LLM, as well as downstream performance. 
    \textbf{(b)} Our method, \textbf{\modelname}, outperforms LLaVA-1.5~\cite{liu2023improvedllava} in both probing and downstream tasks under the same settings.}
    \label{fig:intro}
\end{minipage}
\hfill
\begin{minipage}{0.54\linewidth}
    \includegraphics[width=\linewidth]{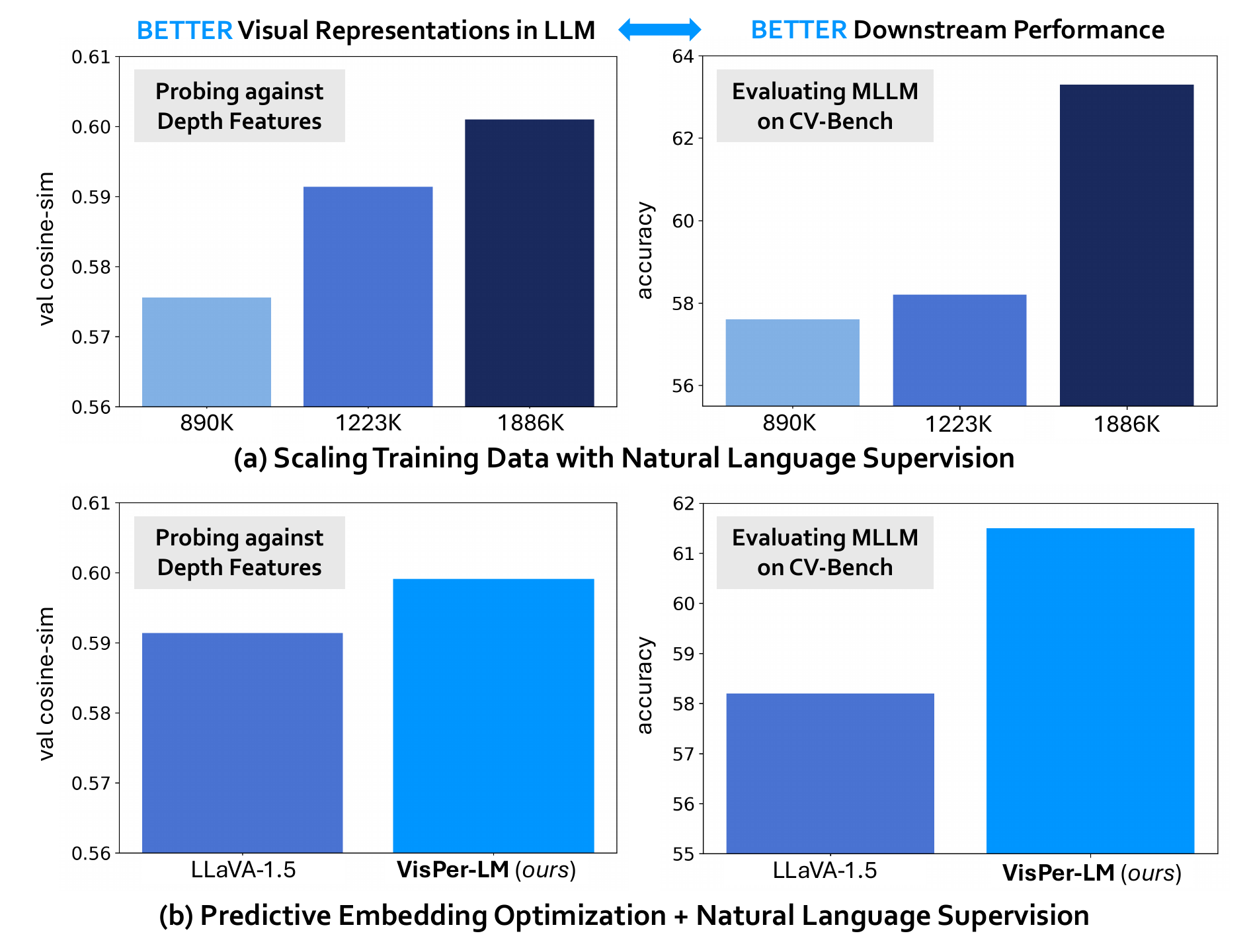}
\end{minipage}%
\vspace{-0.6cm}
\end{figure}

In the last couple of years, Multimodal Large Language Model (MLLM)~\cite{liu2023improvedllava, zhu2023minigpt} development has witnessed rapid growth, owing mainly to the increasing number of powerful LLMs~\cite{abdin2024phi3technicalreporthighly, dubey2024llama3herdmodels, deepseekv2, qwen2} and large-scale datasets~\cite{tong2024cambrian1, wang2023cogvlm, chen2024far}. \changes{A visual reasoning MLLM generally has three main components: vision encoder(s), a projector, and a decoder LLM. Given the presence of the decoder LLM responsible for producing the final outputs, the established practice to train MLLMs is with a next text token prediction objective~\cite{gpt2} on relevant image-text datasets~\cite{llava-one, tong2024cambrian1} with supervision from the ground-truth text}. Specifically, training an MLLM typically involves two primary stages~\cite{liu2023llava}: (i) Pre-Training (PT), training a projector to align the embeddings from the vision encoder(s) to those inside the LLM, and (ii) Instruction Fine-Tuning (IFT), training the projector and LLM on conversation data for downstream instruction following tasks. 
Traditional MLLMs~\cite{liu2023llava} usually use a pre-trained visual encoder like CLIP-ViT~\cite{clip, cherti2023reproducible, liu2023llava} to process the visual inputs.


Although effective on general visual reasoning tasks, the visual encoder's features usually lack the fine-grained visual information required to perform tasks like spatial/depth reasoning~\cite{mmvp,diva,jain2024vcoder}, which are important visual perception abilities. Consequently, several recent works propose simply scaling the number of visual (input) encoders~\cite{tong2024cambrian1, li2024mgm, kar2024brave, jain2024vcoder} to incorporate auxiliary information amicable for improving the model's visual perception abilities. 
\changes{However, existing works often require relatively large amounts of training data~\cite{kar2024brave, tong2024cambrian1} and compute~\cite{nvlm2024, alayrac2022flamingo} for convergence, making these impractical for resource and data limited settings. Moreover, gains from using multiple visual encoders often come at the cost of latency during inference.} In this work, we hypothesize that scaling the number of visual inputs and/or training data improves visual perception performance owing to its positive effect on the visual representations inside the LLM. Therefore, we explore a hidden opportunity to optimize the visual quality of representations inside the LLM directly. To that end, we propose to distill knowledge from a set of expert visual encoders into the LLM's representations during the PT stage through a set of embedding losses (\cref{fig:teaser}\textcolor{cvprblue}{c}). During inference, we only use a single encoder, resulting in a better trade-off between visual understanding performance and efficiency as compared to feeding multiple visual inputs to the LLM (\cref{fig:teaser}\textcolor{cvprblue}{a,b}). 



\changes{We first conduct extensive experiments to establish a relationship between the (M)LLM's visual representation quality and its downstream VQA performance. To that end, we train a probe at each layer inside the LLM (of the MLLM, LLaVA-1.5~\cite{liu2023improvedllava}, in our case due to its wide usage in the academic community) to analyze their quality against expert visual features}. We choose features from visual encoders trained for three tasks as targets: image segmentation~\cite{jain2023oneformer, sam}, depth estimation~\cite{depth_anything_v1, depth_anything_v2}, and image generation~\cite{ldm, unclip}, owing to their well-studied, fundamental nature and the first two being seminal perception abilities~\cite{he2017maskrcnn, eigen2014depth}. \changes{Based on our choice of target features, we refer to benchmarks~\cite{tong2024cambrian1} evaluating the depth and spatial reasoning ability as our target tasks,} i.e., we aim to improve the model at visual perception without harming its general reasoning abilities.  We find that the LLM representations also improve with more data, indicating enhanced visual perception ability and downstream performance, proving the effectiveness of our probing setup (\cref{fig:intro}).




\changes{From our probing experiments, we also observe that the middle LLM layers are optimal for embedding visual information inside the LLM based on the layer-wise representation quality trend.} Consequently, we investigate the effect of optimizing \changes{the intermediate LLM representations against the expert visual features at specific layers}. Inspired by the recent works in embedding predictive architectures for self-supervised learning~\cite{jepa}, we propose \textbf{\modelname} (\textbf{Vis}ual \textbf{Per}ception \textbf{L}anguage \textbf{M}odel) to distill~\cite{kd} knowledge from the expert visual encoders into the LLM's representations \changes{during the pre-training (PT) stage}. Specifically, we optimize an embedding loss between the target feature and the embedding predictor output at the corresponding LLM layer (\cref{fig:teaser}\textcolor{cvprblue}{c}). Note that we still use features from a base encoder as inputs to the LLM. Furthermore, we incorporate a specialized set of tokens, $\langle t \rangle$, enriched with target information into the LLM's input sequence, fostering an implicit visual chain of thought~\cite{goyal2024think, icot} \changes{while enhancing the model's ability to handle spatial and depth reasoning queries}. Our experiments in \cref{sec:exp} illustrate the effectiveness of our approach on various benchmarks while outperforming the baselines. To summarize, our contributions are:





\begin{compactitem}
    \item Through probing existing MLLMs, we \changes{observe a positive correlation between visual representation quality and downstream performance on benchmarks like CV-Bench}. To the best of our knowledge, we are the first to analyze the visual quality of MLLMs' representations.
    \item We present \textbf{\modelname}, an approach to distill knowledge from expert visual encoders into LLM's representations to improve the model's visual (depth/spatial) perception abilities.
    \item We conducted extensive experiments to demonstrate the superiority of \modelname over the corresponding single and multi-encoder baselines on various benchmarks, including boosts up to 8. 7\% and 5. 6\% on the depth and distance task in CV-Bench, respectively.
\end{compactitem}

\section{Related Works}

\begin{figure*}[t!]
\centering
\includegraphics[width=1\linewidth]{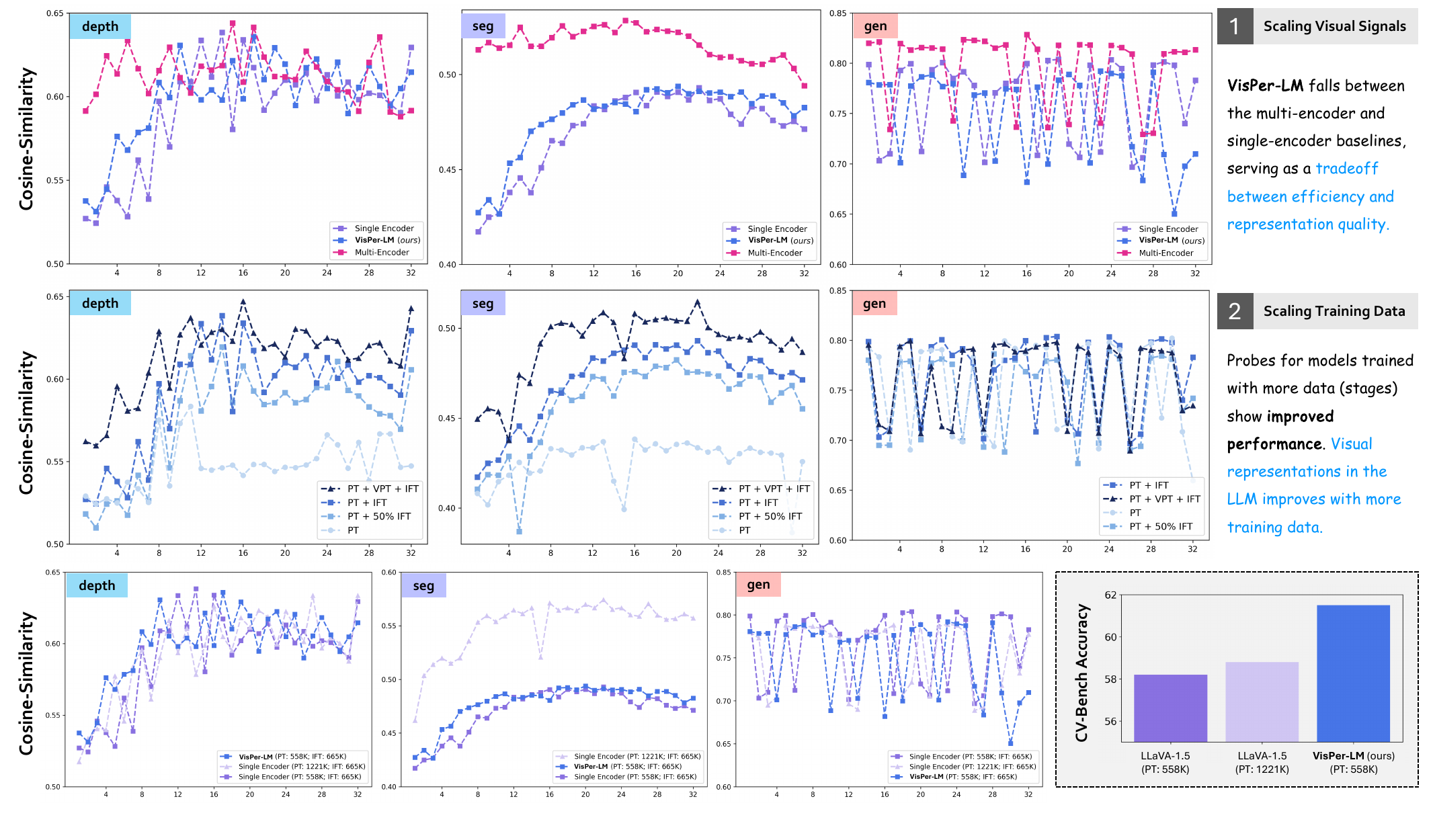} \\
\caption{\textbf{Probing Visual Representation Quality across LLM layers in MLLMs}. \textbf{(1)} As shown in the first row, the multi-encoder baseline has the best probing performance owing to the additional feature inputs. The performance of probes trained on our \modelname falls between the two baselines, demonstrating the effectiveness of our embedding distillation approach in learning an improved projector while only using a single encoder during inference. \textbf{(2)} We observe that the probing performance for single-encoder models trained solely with natural language supervision improves as the training data of the base MLLM increases, indicating that the LLM improves its visual representations of the world with more training data. In the last row, we observe that our \modelname (base setting) outperforms LLaVA-1.5 trained with more data during the PT stage, \changes{demonstrating the effectiveness of our approach with limited (data/compute) resources}.}
\vspace{-0.4cm}
\label{fig:teaser_plots}
\end{figure*}

 
\subsection{MLLMs for Visual Reasoning}

Contemporary MLLMs have three primary components: vision encoder(s), projector, and LLM. 
One line of work~\cite{zhu2023minigpt, llava-one, liu2023improvedllava, li2024cumo, liu2023llava} uses a single pretrained vision encoder~\cite{clip, vit, convnext, cherti2023reproducible, siglip} and trains a projector, like an MLP~\cite{liu2023improvedllava} or QFormer~\cite{instructblip} to align the visual features with the LLM. A few recent approaches~\cite{chameleon, diao2024EVE} attempt to develop native MLLMs by directly feeding image patches into the LLM without using any visual encoder. Some works~\cite{alayrac2022flamingo, nvlm2024} train cross-attention modules inside the LLM, modifying the LLM architecture and often requiring millions of training data samples.
Another line of work explores using either multiple encoder features~\cite{tong2024cambrian1, mgllava, shi2024eagle, li2024mgm} or multiple visual modalities~\cite{jain2024vcoder, liu2024prismer, cheng2024spatialrgpt} as inputs to the LLM for improved visual (spatial) reasoning performance. Our approach uses a single-base vision encoder while distilling information from expert visual encoders into the LLM's representations.

\subsection{Perception Probing in Foundation Models}

OthelloGPT~\cite{othello} probed the features from a GPT-2~\cite{gpt2} trained on sequences from a board game, Othello, and found that the probes were able to learn the board state, indicating the ability of sequence models to learn spatial world representations. A recent work probes the features from foundational vision encoders~\cite{banani2024probing3dawarenessvisual} for 3D tasks. In our work, we probe the representation from the LLM layers against visual features from expert perception models. Furthermore, we establish a relationship between visual representation quality inside the LLM and downstream VQA performance.

\subsection{Self-Supervised Learning}

Distilling information~\cite{kd} from a target encoder into a source encoder is a well-established technique to improve the source encoder's embeddings for a downstream task~\cite{simclr, chen2020exploring, grill2020bootstrap}. Recently, I-JEPA~\cite{jepa} proposed an embedding predictive architecture to improve representations inside a source encoder by comparing the target encoder features and mapped source encoder features with a trained predictor. A concurrent work, REPA~\cite{yu2024repa} improved DiTs~\cite{Peebles2022DiT} at image generation by distilling information from DINOv2~\cite{oquab2023dinov2}. Recently, a few works~\cite{radio, diva} first distill information from target encoders into a single model and then leverage the resulting model as the vision encoder in an MLLM, \changes{requiring relatively longer and more training data for distillation}. Unlike previous works, we distill information from multiple vision encoders into the LLM embedding space during the PT stage, resulting in a more vision-centric training approach \changes{without any extra training data.}

\section{Visually Probing LLM's Embeddings}
\label{sec:probe}

\begin{figure}[t!]
\centering
\includegraphics[width=1\linewidth]{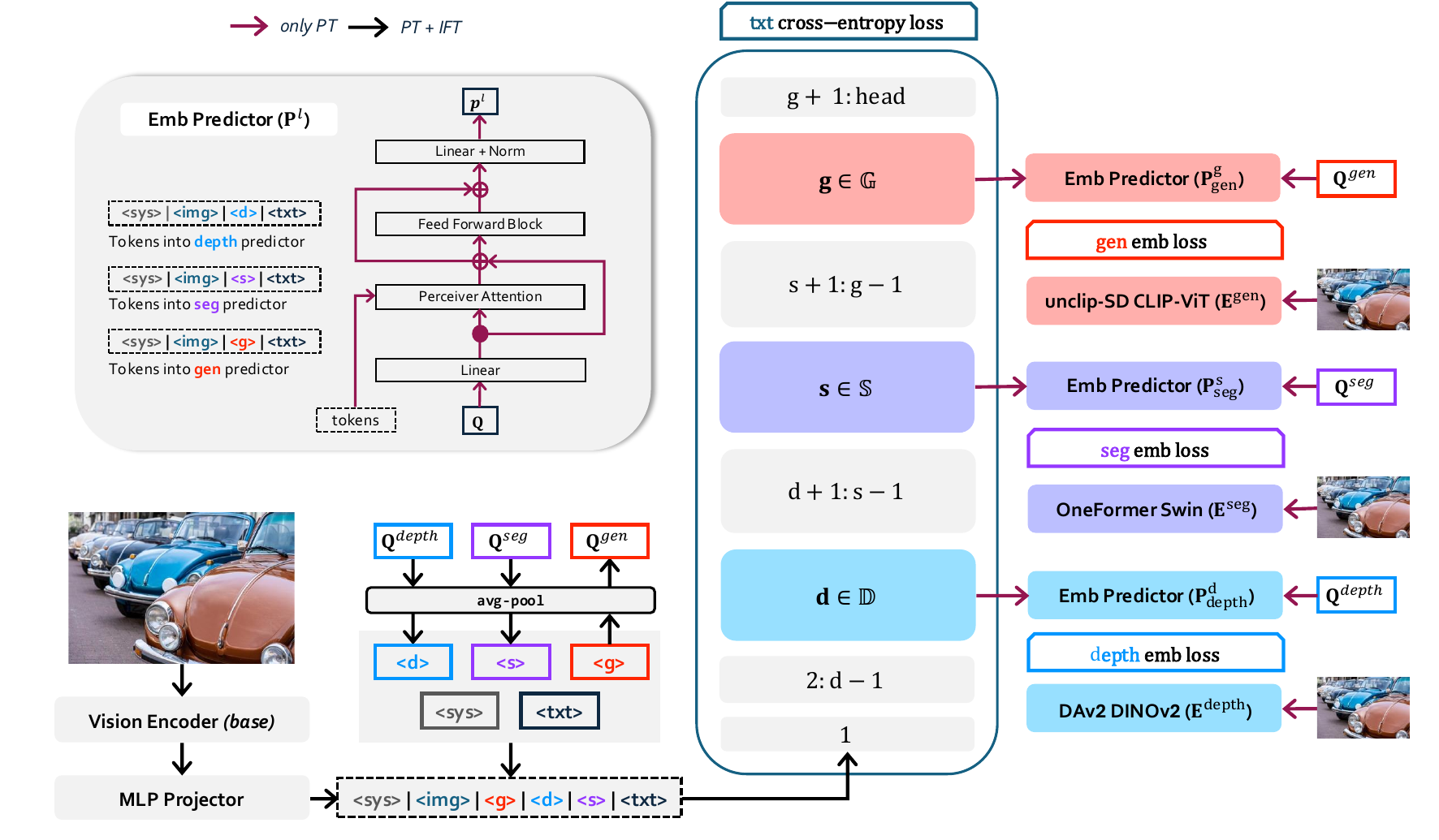} \\
\vspace{-0.1cm}
\caption{\textbf{Architecture for \modelname}. During Pre-Training (PT), we optimize an embedding loss at specific layers for each target encoder: layers $d \in \mathbb{D}$, $s \in \mathbb{S}$, and $g \in \mathbb{G}$ for the depth, segmentation, and generation tasks, respectively. We use a resampler-based embedding predictor~\cite{perceiver}, denoted as $\mathbf{P}^{l}_{\{\text{task}\}}$ at each layer $l$, to output predictions. Each predictor takes in two inputs: a set of learnable queries ($\mathbf{Q}^{{\text{task}}}$) and the token sequence from layer $l$, with special tokens for other tasks omitted. The task tokens are derived from the corresponding embedding predictor's learnable queries.  During IFT, we train with only the next-token prediction objective while keeping the special tokens frozen to not affect their nature as we found it to perform empirically better in \cref{sec:exp}.}
\vspace{-0.5cm}
\label{fig:sherlock_arch}
\end{figure}

In this section, we systematically analyze the quality of LLM representations within the MLLM through probing with the expert visual features as the targets. We define high-quality representations as those that map easily to the target feature space, demonstrated by a high cosine similarity to the corresponding expert encoder’s features.
We choose target encoders from models trained for three visual tasks: image segmentation,  generation, and depth estimation, guided by the mentioned tasks' fundamental and interpretable nature, i.e., visualizing the representations' quality through the respective decoders. We use the encoder outputs from Swin-L OneFormer~\cite{jain2023oneformer, swin-T}, DINOv2-L Depth Anything-v2~\cite{vit, depth_anything_v2}, and CLIP-ViT-L unCLIP-SD-2.1~\cite{unclip, clip} as the segmentation ($\mathbf{E}^\text{seg}$), depth ($\mathbf{E}^\text{depth}$) and generation ($\mathbf{E}^\text{gen}$) as target probe features, respectively.


\noindent
\textbf{Probing Setup.} We train a single-layer Perceiver Resampler~\cite{perceiver} as the probe head at every LLM layer for each of the three target features. We use a resampler probe head (similar to the Emb Predictor shown in \cref{fig:sherlock_arch}) to accommodate the different sequence lengths of the target features and representations inside the LLM. \changes{For each LLM layer, we input learnable queries into the probe head while using the layer's hidden states (for all tokens) as keys for cross-attention inside the resampler block}. During probing, we set the learning rate as $1e^{-3}$ and minimize the smooth-L1-loss objective with a batch size of 256. We train the probes for two epochs on the 118k images from the COCO-train2017~\cite{coco} set with the text query: \textit{``Describe the image in two lines."}. \changes{We find a positive correlation of 0.98 between the depth probing performance and CV-Bench accuracy in \cref{tab:corr}, proving the effectiveness of our probing setup}.
We analyze the following LLaVA-1.5-based models with Llama-3-8b~\cite{dubey2024llama3herdmodels} as the LLM and CLIP-ConvNeXT-XXL~\cite{cherti2023reproducible} as the visual encoder:

\begin{compactitem}
    \item \underline{Single-Encoder} MLLM with features from $\mathbf{E}^\text{base}$ passed through an MLP into the LLM.
    \item \underline{Multi-Encoder} MLLM with features from $\mathbf{E}^\text{base}$, $\mathbf{E}^\text{gen}$, $\mathbf{E}^\text{depth}$, and $\mathbf{E}^\text{seg}$ concatenated in the feature dimension (in order) and passed through an MLP into the LLM.
    \item \textbf{\modelname} with features from $\mathbf{E}^\text{base}$ passed through an MLP into the LLM and information from $\mathbf{E}^\text{gen}$, $\mathbf{E}^\text{depth}$, and $\mathbf{E}^\text{seg}$ distilled into the LLM representations.
\end{compactitem}

We compute the cosine similarity between the probe outputs and the corresponding target features over the 5k images from the COCO-val2017~\cite{coco} set to get the probing performance for evaluation. 

\noindent
\textbf{Layerwise Trend.}
Upon evaluating the probes in the single encoder baseline, we observe that the \hl{middle (12–24) layer probes show the best representation quality for the depth and seg probing tasks, with an upward trend in quality in the initial layers and a downward trend in the deeper layers}, as shown in \cref{fig:teaser_plots}. \changes{We attribute the observed trend to the fact that the middle layer representations contribute the most to the (visual) reasoning (primarily spatial/depth) in LLMs~\cite{kaduri2024_vision_of_vlms, kaplan2024tokenswords, skean2024layer_rep}. It is an important finding when deciding the position of the embedding losses, since optimizing the middle layer representations should be optimal.} Surprisingly, the probes trained to predict the features from $\mathbf{E}^\text{gen}$ have fairly high cosine similarity (greater than 0.7) for all layers. We attribute the mentioned phenomenon to the choice of $\mathbf{E}^\text{gen}$ that already has language-aligned features, unlike $\mathbf{E}^\text{depth}$ and $\mathbf{E}^\text{seg}$.

\noindent
\textbf{Visual Encoding Approach.} As shown in the first row of \cref{fig:teaser_plots}, the probes for multi-encoder MLLM expectedly learn better representations than the probes for the single-encoder MLLM. Interestingly, for the former, the depth and seg probe performance remain at the same level till layer 20 and then follow a downward trend, indicating the possibility of the deeper layer representations becoming text-rich in deeper layers~\cite{lin2024boosting}. We do not observe a downward trend for gen probing, owing to the text-aligned target gen features. We observe that the \hlgreen{VisPer-LM probes fall between the two baselines, serving as a good trade-off between efficiency and visual representation quality.}

\noindent
\textbf{Training Data.} We study the effect of scaling training data on the visual representation quality inside the single-encoder MLLM in the second row of \cref{fig:teaser_plots}. We analyze four models: (i) PT: model after the PT stage, (ii) PT + 50\% IFT: model with complete PT and trained on 50\% IFT data, (iii) PT + IFT: model with complete PT and IFT, and (iv) PT + VPT + IFT: model with an extra training stage on ALLaVA-Caption~\cite{chen2024allava} data, during which the whole model is trained. We observe that \hlred{with an increase in the amount of training data for the probed model, the probes show a gradual improvement}, indicating that the representations of the visual world inside MLLMs and, therefore, downstream performance improve with just natural language supervision on more data (\cref{fig:intro}\textcolor{cvprblue}{a})!

For experimental completeness, we also report probing for the single-encoder LLaVA-1.5 trained with additional ALLaVA-Caption~\cite{chen2024allava} data during the PT stage. As illustrated in the third row of \cref{fig:teaser_plots}, while the inclusion of additional PT data improves LLaVA-1.5's performance on CV-Bench, \hlred{our VisPer-LM achieves superior results with lesser PT data, highlighting the effectiveness of our approach with limited data}. Notably, there is a significant improvement in representation quality for the seg probing task with the use of ALLaVA-Caption during PT, which we attribute to the dataset's detailed captions fostering enhanced semantic alignment~\cite{chen2024allava}. Please refer to \cref{tab:extra_pt} for more results.

\section{Embedding Visual Information into LLM}
\label{sec:method}


The current trend in developing MLLMs for visual reasoning is gluing vision encoder(s) to a decoder LLM with a projector and training the model to minimize the cross-entropy loss for classification over the LLM's vocabulary for next token prediction (NTP). In our attempt to add a vision perspective to training MLLMs while only using a single visual encoder during inference, we aim to optimize a predictive visual embedding objective for the intermediate LLM representations with the standard NTP objective during the PT stage. In other words, we distill auxiliary visual (task) information into the LLM's intermediate representations rather than feeding the visual features into the LLM from the expert encoders. We hypothesize that such an approach leads to a better projector initialization (\cref{tab:prob_ref}) for the IFT stage. Our hypothesis is validated in \cref{fig:teaser_plots}, as probes for our \modelname significantly outperform the single encoder baseline, especially in the initial layers. During IFT, we train the MLLM using the NTP objective, without the embedding predictors.





\subsection{Multi-Encoder Feature Distillation}



Given a set of target visual encoders, $\mathbf{E}^\text{depth}$, $\mathbf{E}^\text{seg}$, and $\mathbf{E}^\text{gen}$, we distill information from their feature outputs into the LLM's representation space by minimizing embedding losses between predictor outputs and target features at certain layers. Since the token sequence inside the LLM has a different length as compared to the target representations, we use a single-layer Perceiver Resampler~\cite{perceiver} as the embedding predictor ($\mathbf{P}^l_\text{\{task\}}$) that takes as inputs: learnable latent queries ($\mathbf{Q}^\text{\{task\}}$) along with the embeddings from layer $l$ of the LLM and outputs a prediction, $\mathbf{p}^l$. We set the number of queries in $\mathbf{Q}^\text{\{task\}}$ such that it matches the number of tokens in the corresponding $\mathbf{E}^\text{\{task\}}$ features, \emph{i.e.}, $576$, $576$, and $1$, for $\mathbf{Q}^{depth}$, $\mathbf{Q}^{seg}$, and $\mathbf{Q}^{gen}$, respectively.

To amplify the effect of the contextual information about the target task (spatial/depth reasoning) inside the LLM~\cite{chen2023minigptv2,goyal2024think}, we use a set of special $\mathbf{N}$ task tokens, $\langle t \rangle $. We append $\langle t \rangle $ to the image tokens in the LLM's input sequence. Specifically, we average pool $\mathbf{Q}^{depth}$ and $\mathbf{Q}^{seg}$ into $\mathbf{N}$ number of tokens to obtain the \bl{$\langle d \rangle $} and \epp{$\langle s \rangle $} tokens, respectively. Since the number of target tokens for generation features is only one, we average pool \re{$\langle g \rangle $} to obtain $\mathbf{Q}^{gen}$. During PT, we only train the MLP projector, embedding predictors, and the special tokens $\langle t \rangle $. We extract the tokens corresponding to the system prompt, input image, the corresponding $\langle t \rangle $, and the text query from a layer's output sequence as the input keys to the embedding predictor, 
as shown in \cref{fig:sherlock_arch}.



\subsection{Predictive Embedding Optimization}


At the core of our approach is indirectly optimizing the projector during the PT stage by minimizing an embedding loss for each target representation at specific layers. As shown in \cref{fig:sherlock_arch}, we feed the outputs from the $\mathbf{d}\in \mathbb{D}$, $\mathbf{s}\in \mathbb{S}$, and $\mathbf{g}\in \mathbb{G}$ layers of the LLM into the corresponding embedding predictor to obtain the predictions for computing an embedding loss ($\mathcal{L}_\text{emb}$), which is a weighted sum of Smooth-L1-Loss~\cite{fast-rcnn} and contrastive (InfoNCE) loss~\cite{oord2018representation}. We denote the sets of layers for embedding distillation from $\mathbf{E}^\text{depth}$, $\mathbf{E}^\text{seg}$, and $\mathbf{E}^\text{gen}$ as $\mathbb{D}$, $\mathbb{S}$, and $\mathbb{G}$, respectively. The final embedding loss for each target feature is a sum of losses over all corresponding layers. We compute the final loss during PT as the sum of the NTP objective and embedding losses, as shown in \cref{eq:loss}.

We denote $\mathbf{p}^l$ and $\mathbf{t}$ as the embedding predictor ($\mathbf{P}^l_{\{\text{task}\}}$) outputs at layer $l$ and the target task features, respectively. $\mathbf{p}^l_{i}$ denotes the $i^\text{th}$ element in a batch of embeddings with a global batch size of $B$ aggregated over all GPUs. We denote $\tau$ (initialized to $2.0$) as the learnable scaling factor for the contrastive loss. The weights for smooth-L1-and contrastive losses are $\lambda_{\text{sL1}} = 1$ and $\lambda_{\text{contrastive}} = 0.3$, respectively, at all selected layers. We select these values to ensure their magnitudes are comparable and their weighted sum aligns in scale with $\mathcal{L}_\text{NTP}$ at convergence to maintain training stability. We set $\lambda_\text{depth}$, $\lambda_\text{seg}$, and $\lambda_\text{gen}$ as 0.5.

\vspace{-0.3cm}
\begin{equation}
\mathcal{L}^l_{\text{sL1}}(\mathbf{p}^l, \mathbf{t}) = 0.5 \cdot (\mathbf{p}^l - \mathbf{t})^2 \texttt{  if  } |\mathbf{p}^l - \mathbf{t}| < 1 \texttt{  else  } |\mathbf{p}^l - \mathbf{t}| - 0.5
\end{equation}
\vspace{-0.8cm}

\begin{equation}
\mathcal{L}^l_{\text{contrastive}} = -\log \frac{\exp(\text{sim}(\mathbf{p}^l_i, \mathbf{t}_i) / \tau)}{\sum_{j=1}^{B} \exp(\text{sim}(\mathbf{p}^l_i, \mathbf{t}_j) / \tau)}
\end{equation}
\vspace{-0.6cm}

\begin{equation}
\mathcal{L}^{\mathbb{D/S/G}}_{\text{emb}} = \sum_{l \in \mathbb{D/S/G}} \left( \lambda_{\text{sL1}}\mathcal{L}^l_{\text{sL1}} + \lambda_{\text{contrastive}} \mathcal{L}^l_{\text{contrastive}} \right)
\end{equation}
\vspace{-0.8cm}

\begin{equation}
\mathcal{L}_\text{PT} = \mathcal{L}_\text{NTP} + \lambda_\text{depth}\mathcal{L}^{\mathbb{D}}_{\text{emb}} + \lambda_\text{seg}\mathcal{L}^{\mathbb{S}}_{\text{emb}} + \lambda_\text{gen}\mathcal{L}^{\mathbb{G}}_{\text{emb}}
\label{eq:loss}
\end{equation}

\noindent
\textbf{Other Architecture Details.}
As shown in \cref{fig:sherlock_arch}, we use DINOv2-L~\cite{oquab2023dinov2} from Depth Anything v2~\cite{depth_anything_v2} as $\mathbf{E}^\text{depth}$, Swin-L~\cite{swin-T} from OneFormer~\cite{jain2023oneformer} (trained on COCO-train2017~\cite{coco}) as $\mathbf{E}^\text{seg}$, and CLIP-ViT-L~\cite{clip} from unCLIP-SD-2.1~\cite{unclip} as $\mathbf{E}^\text{gen}$. In the input sequence to the LLM, we append the gen, depth, and seg tokens in that order to the image tokens. Therefore, given an image-text pair, the input token arrangement is 
\{$\langle sys \rangle | \langle img \rangle | \re{\langle g \rangle} | \bl{\langle d \rangle} | \epp{\langle s \rangle} |  \langle txt \rangle $\}, where $\langle sys \rangle$, $\langle img \rangle$, and $\langle txt \rangle$ denote the tokens corresponding to the system prompt, image, and text query, respectively.
The target feature dimensions from $\mathbf{E}^\text{gen}$, $\mathbf{E}^\text{depth}$, and $\mathbf{E}^\text{seg}$ are $(1, 1024)$, $(576, 1024)$, and $(576, 1536)$, respectively, corresponding to their final layer outputs.
\section{Experiments}
\label{sec:exp}

\begin{table}[t!]
  \centering
  \caption{\textbf{Comparisons to Baselines.} Our \modelname outperforms the single and multi-encoder LLaVA-1.5~\cite{liu2023improvedllava} by up to 2.5\% and 0.9\% on average across various benchmarks, respectively. The best numbers are set in \textbf{bold} for every base-encoder and decoder LLM combination.}
  \vspace{0.1cm}
  \resizebox{1.\linewidth}{!}{
  \begin{tabular}{llcccc|ccc|c}

 & & \multicolumn{4}{c|}{\textbf{CV-Bench}} &  \multicolumn{3}{c|}{\textbf{General}} & \\
\midrule

\textbf{Method} & \textbf{Encoder} & \textbf{Count}$^\text{2D}$ & \textbf{Depth}$^\text{3D}$ & \textbf{Relation}$^\text{2D}$ & \textbf{Distance}$^\text{3D}$ & \textbf{MMStar} & \textbf{RWQA} & \textbf{OK-VQA} & \textbf{Avg} \\
\toprule

\multicolumn{6}{l}{\textit{Phi3-4k-mini}} \\
\midrule

LLaVA-1.5 & CLIP-ViT-L &  \textbf{52.4} & 67.2 & 75.2 & 56.3 & \textbf{36.5} & 57.1 & \textbf{56.7} & 57.3 \\

\rowcolor{azure!10}
\textbf{\modelname} (ours) & CLIP-ViT-L &  \textbf{52.4} & \textbf{68.7} & \textbf{76.0} & \textbf{56.7} & 36.0 & \textbf{58.0} & 56.4 & \textbf{57.7} \\

\midrule

LLaVA-1.5 & CLIP-ConvNeXT-XXL & \textbf{51.8} & 70.8 & 74.0 & 55.3 & 36.4 & 58.0 & 55.9 & 57.4 \\

\rowcolor{azure!10}
\textbf{\modelname} (ours) & CLIP-ConvNeXT-XXL &  49.4 & \textbf{72.5} & \textbf{77.2} & \textbf{60.3} & \textbf{38.4} & \textbf{58.4} & \textbf{56.5} & \textbf{58.9} \\

\midrule
\multicolumn{6}{l}{\textit{Llama3-8b}} \\
\midrule


LLaVA-1.5 & CLIP-ViT-L & 50.4 & 73.3 & 64.9 & 48.7 & 38.8 & 57.8 & {56.9} & 55.1 \\

\rowcolor{gray!15}
LLaVA-1.5 (feat concat.) & CLIP-ViT-L + $\mathbf{E}^\text{depth}$ + $\mathbf{E}^\text{seg}$ + $\mathbf{E}^\text{gen}$ & 45.3 & 75.5 & \textbf{70.9} & \textbf{54.3} & 36.1 & 57.5 & 58.3 & 56.8 \\

\rowcolor{gray!15}
LLaVA-1.5 (token concat.) & CLIP-ViT-L + $\mathbf{E}^\text{depth}$ + $\mathbf{E}^\text{seg}$ + $\mathbf{E}^\text{gen}$ & 45.9 & \textbf{75.7} & 68.9 & 52.7 & 37.8 & 56.5 & \textbf{59.3} & 56.7 \\

\rowcolor{azure!10}
\textbf{\modelname} (ours) & CLIP-ViT-L & \textbf{51.3} & {74.2} & {69.4} & \textbf{54.3} & \textbf{39.5} & \textbf{57.9} & 56.6 & \textbf{57.6} \\

\midrule

\rowcolor{gray!15}
Cambrian-1~\cite{tong2024cambrian1} & \begin{tabular}{@{}l@{}} CLIP-ConvNeXT-XXL + CLIP-ViT-L + \\ SigLIP-SO-400M~\cite{siglip} + DINOv2-G~\cite{oquab2023dinov2} \end{tabular} & 57.2 & 65.0 & 63.2 & 50.1 & 34.5 & 53.2 & --- & ---  \\
\midrule
\rowcolor{gray!15}
LLaVA-1.5 & RADIO-ViT-L~\cite{radio} & 56.4 & 64.5 & 65.8 & 50.1 & 36.6 & 55.2 & 57.4 & 55.1 \\

LLaVA-1.5 & CLIP-ConvNeXT-XXL & 54.1 & 62.8 & \textbf{69.5} & 49.8 & 37.4 & \textbf{57.5} & 56.3 & 55.3 \\

\rowcolor{azure!10}
\textbf{\modelname} (ours) & CLIP-ConvNeXT-XXL & \textbf{57.4} & \textbf{71.5} & 66.8 & \textbf{52.8} & \textbf{38.5} & 55.0 & \textbf{59.0} & \textbf{57.3} \\

\bottomrule
\end{tabular}

}
    \label{tab:res}
    \vspace{-0.2cm}
\end{table}

\begin{table}[t!]
  \centering
  \caption{\textbf{Results on Additional Benchmarks.} Our \modelname outperforms LLaVA-1.5 on classical benchmarks like POPE~\cite{pope} and GQA~\cite{gqa} showing its intact general reasoning ability.}
  \vspace{0.2cm}
  \resizebox{0.7\linewidth}{!}{
  \begin{tabular}{ll|cccc|c}
\textbf{Method} & \textbf{Encoder} & \textbf{POPE} & \textbf{GQA} & \textbf{MMMU$^{\text{val}}$} & \textbf{VizWiz$^{\text{val}}$} & \textbf{Avg} \\
\toprule

LLaVA-1.5 & RADIO-ViT-L~\cite{radio} & 86.3 & 62.8 & 36.9 & 50.5 & 59.1 \\

LLaVA-1.5 & CLIP-ViT-L & 85.9 & 63.5 & 38.6 & 50.6 & 59.7 \\

\rowcolor{azure!10}
\textbf{\modelname} (ours)  &  CLIP-ViT-L & \textbf{86.4} & \textbf{63.7} & \textbf{38.7} & \textbf{54.0} & \textbf{60.7} \\

\end{tabular}}
    \label{tab:extra-bench}
    \vspace{-0.2cm}
\end{table}

\begin{table}[t!]
  \centering
  \caption{\textbf{Scalability with VPT (more training data).} \modelname outperforms LLaVA-1.5 on average across different CV-Bench tasks. We use CLIP-ConvNeXT-XXL as the base encoder.}
  \vspace{0.2cm}
  \resizebox{0.7\linewidth}{!}{
  \begin{tabular}{ll|cccc|c}
\textbf{Method} & \textbf{LLM} & \textbf{Count}$^\text{2D}$ & \textbf{Depth}$^\text{3D}$ & \textbf{Relation}$^\text{2D}$ & \textbf{Distance}$^\text{3D}$ & \textbf{Overall} \\
\toprule

LLaVA-1.5 & Phi3-4k-mini & 49.7 & 70.0 & 72.6 & \textbf{58.7} & 61.8 \\

\rowcolor{azure!10}
\textbf{\modelname} (ours) & Phi3-4k-mini & \textbf{53.7} & \textbf{72.0} & \textbf{73.1} & 58.5 & \textbf{63.4} \\

\midrule

LLaVA-1.5 & Llama3-8b & 56.3 & \textbf{76.8} & \textbf{73.1} & 50.3 & 63.3 \\
\rowcolor{azure!10}
\textbf{\modelname} (ours) & Llama3-8b & \textbf{60.0} & 75.0 & 70.8 & \textbf{55.2} & \textbf{64.6} \\

\bottomrule
\end{tabular}}
  \vspace{-0.2cm}
    \label{tab:abl_vpt}
\end{table}

In this section, we first provide a comprehensive comparison of our method's performance to that of the base MLLM, LLaVA-1.5~\cite{liu2023improvedllava} across different base vision encoder and decoder LLM choices in \cref{tab:res}. Next, we provide experimental results with a 2.5-stage training strategy, composed of an extra Visual Pre-Training (VPT) stage that involves training on the ALLaVA-Caption-663K~\cite{chen2024allava} dataset to demonstrate the scalability of our approach to more data. Lastly, we methodically study various design factors, including the optimal choice of layers for embedding additional visual information and the number and nature of special tokens ($\langle t \rangle $) through a series of ablations.

\subsection{Implementation Details}
\label{subsec:impl}

\noindent
\textbf{Training.} During the PT stage, we use the LLaVA-558K~\cite{liu2023improvedllava} dataset to train our model for an epoch with $lr$ of $1e^{-3}$. We only train the (MLP) projector, the embedding predictors, and the special tokens ({$\langle t \rangle $}).
During the IFT stage, we use the LLaVA-665K~\cite{liu2023improvedllava} dataset and train the projector and LLM for one epoch with an $lr$ of $2e^{-5}$ with the vision encoder and $\langle t \rangle $ kept frozen. When using VPT, we leverage the ALLaVA-Caption-663K~\cite{chen2024allava} dataset to train the whole model (except $\langle t \rangle $) for one epoch with an $lr$ of $2e^{-5}$. Unless mentioned otherwise, we report results with models trained with PT and IFT stages. During the PT stage, we train our model with the NTP objective and the embedding prediction objectives. During the VPT and IFT stages, we solely use the NTP objective.

We train all our models on 16 AMD 192G-MI300X GPUs with a batch size of 256 during PT and 128 during IFT and VPT. We use CLIP-ViT-L~\cite{clip} and Llama3-8b~\cite{dubey2024llama3herdmodels} as the base vision encoder and decoder LLM unless mentioned otherwise. By default, we set $\mathbf{N}$, $\mathbb{D}$, $\mathbb{S}$, and $\mathbb{G}$ to 8, $\{8,20\}$, $\{10,18\}$, and $\{12,20\}$, respectively. For other hyperparameters, we follow LLaVA-1.5~\cite{liu2023improvedllava}. 


\noindent
\textbf{Evaluation.} We primarily evaluate \modelname for vision-centric abilities on CV-Bench~\cite{tong2024cambrian1} and report results for all four tasks in CV-Bench: \textit{Count} (2D), \textit{Relation} (2D), \textit{Depth} (3D), and \textit{Distance} (3D). For ablations, we report the average accuracy over the 2D (\textit{count} and \textit{relation}) and 3D (\textit{depth} and \textit{distance}) task categories. We also evaluate our models' general visual reasoning capabilities on the MMStar~\cite{mmstar}, RWQA~\cite{xai_grok_2024}, and OK-VQA~\cite{okvqa} benchmarks. \changes{We select MMStar as our primary benchmark for general visual reasoning, as it addresses the well-known limitations~\cite{mmstar} of existing benchmarks~\cite{MMBench, lu2022learn, yue2023mmmu, li2023seed} through careful filtering. Nonetheless, we report results on several classical general reasoning benchmarks~\cite{gqa,pope,vizwiz,yue2023mmmu} in \cref{tab:extra-bench} for experimental completeness.}

\begin{table}[t!]
  \centering
  \caption{\textbf{Comparison on the Perception tasks using BLINK~\cite{fu2024blink} benchmark.} Our \modelname improves significantly over LLaVA-1.5~\cite{liu2023improvedllava} on depth and relation reasoning, our target perception tasks.}
  \vspace{0.1cm}
  \resizebox{0.8\linewidth}{!}{
  \begin{tabular}{l|c|ccc}
\textbf{Method} & \textbf{Avg} & \textbf{Spatial Relation} & \textbf{Relative Depth} & \textbf{Count}  \\
\toprule

LLaVA-1.5 & 55.6 & 72.0 & 51.4 & 43.3 \\

RADIO-L LLaVA-1.5 & 55.8 & 65.7 & \textbf{52.4} & \textbf{49.2} \\

Cambrian-1 & 56.9 & 74.1 & 51.6 & 45.0 \\

\rowcolor{azure!10}
\textbf{\modelname} (ours) & \textbf{58.8} & \textbf{75.5} & 51.6 & \textbf{49.2} \\
\bottomrule
\end{tabular}
}
  \vspace{-0.2cm}
    \label{tab:blink}
\end{table}

\begin{table}[t!]
  \centering
  \caption{\textbf{Ablations on Layer sets for $\mathcal{L}_\text{emb}$.} Setting $\mathbb{D}$=$\{8,20\}$, $\mathbb{S}$=$\{10,18\}$, and $\mathbb{G}$=$\{12,20\}$ performs the best. Following the findings in \cref{sec:probe}, we only experiment with middle layers.}
  \vspace{0.1cm}
  \resizebox{0.8\linewidth}{!}{
  \begin{tabular}{lll|ccc|c}
$\mathbb{D}$ & $\mathbb{S}$ & $\mathbb{G}$ & \textbf{CV-Bench}$^{2D}$ & \textbf{CV-Bench}$^{3D}$ & \textbf{MMStar} & \textbf{Avg} \\
\toprule


\{20\} & \{18\} & \{20\} & 57.6 & 60.8 & 38.8 & 52.4 \\


\rowcolor{azure!10}
\{8;20\} & \{10;18\} & \{12;20\} & \textbf{58.6} & \textbf{64.2} & 39.5 & \textbf{54.1} \\

\{18;20\} & \{18;20\} & \{16;20\} & 55.8 & 59.5 & \textbf{40.8} & 52.0 \\

\{16;18;20\} & \{16;18;20\} & \{16;18;20\} & 56.8 & 61.3 & 37.0 & 51.7 \\

\bottomrule
\end{tabular}
}
  \vspace{-0.4cm}
    \label{tab:abl-layers}
\end{table}

\begin{table}[t!]
  \centering
  \caption{\textbf{Training stages for $\mathcal{L}_\text{emb}$.} Using the embedding losses only during PT is optimal.}
  \vspace{0.1cm}
  \resizebox{0.6\linewidth}{!}{
  \begin{tabular}{cc|cccc|c}
\textbf{PT} & \textbf{IFT} & \textbf{CV-Bench}$^{2D}$ & \textbf{CV-Bench}$^{3D}$ & \textbf{MMStar} & \textbf{RWQA} & \textbf{Avg} \\
\toprule

\multicolumn{2}{c|}{single-encoder} & 56.0 & 61.0 & 38.8 & 57.8 & 53.4 \\
\midrule

 & & 57.7 & 62.9 & 38.8 & 57.5 & 54.2 \\

\rowcolor{azure!10}
\checkmark & & {58.6} & \textbf{64.2} & \textbf{39.5} & \textbf{57.9} & \textbf{55.1} \\

\checkmark & \checkmark & \textbf{59.1} & 58.3 & 38.3 & 56.2 & 53.0 \\

\bottomrule
\end{tabular}}
  \vspace{-0.2cm}
    \label{tab:abl_stage}
\end{table}

\subsection{Main Results}

As shown in \cref{tab:res}, our \modelname outperforms the single encoder baseline, i.e., LLaVA-1.5~\cite{liu2023improvedllava} across different base encoder~\cite{cherti2023reproducible, clip} and decoder LLM~\cite{dubey2024llama3herdmodels, abdin2024phi3technicalreporthighly} combinations. Specifically, Llama3-8b~\cite{dubey2024llama3herdmodels} based \modelname outperforms LLaVA-1.5 by \textbf{5.6\%} and \textbf{4.5\%} on the \textit{Distance} and \textit{Relation} task, respectively for the CLIP-ViT-L base encoder and by \textbf{8.7\%} on the \textit{Depth} task for the CLIP-ConvNext-XXL base encoder. Furthermore, we compare our approach to the corresponding two variants of multi-encoder baselines: (i) \textit{feat concat.}: features from all encoders are concatenated along the feature dimension and passed into a single MLP; and (ii) \textit{token concat.}: features from all encoders are first passed through separate MLPs and then concatenated along with token dimension after average pooling the sequence outputs from $\mathbf{E}^\text{depth}$ and $\mathbf{E}^\text{seg}$ into eight tokens each. As shown in \cref{tab:res}, our \modelname outperforms the multi-encoder baselines on average, showing the effectiveness of our approach while using a single encoder during inference. \changes{Note that the multi-encoder baselines are better than \modelname on the Depth task in CV-Bench owing to their superior depth representation quality as observed in \cref{sec:probe}.}

We also compare our \modelname with a RADIO-ViT-L~\cite{radio} and Llama3-8b-based LLaVA-1.5. Unlike our approach, which distills information directly into the LLM representations, RADIO distills knowledge from multiple expert encoders into a single student encoder. As shown in \cref{tab:res}, our \modelname outperforms the RADIO-ViT-L-based LLaVA-1.5, underscoring the effectiveness of our approach over encoder distillation for developing MLLMs. Note that we train the RADIO-ViT-L-based model using the same setting as the single encoder baseline.

Additionally, we train a Llama3-8b based Cambrian-1~\cite{tong2024cambrian1} model (with a batch size of 512, following Tong \textit{et al.}~\cite{tong2024cambrian1}) using the LLaVA-558K (PT) plus LLaVA-665K (IFT) dataset mix. As shown in \cref{tab:res}, the Cambrian-1 model not only underperforms as compared to our \modelname but also LLaVA-1.5, underscoring the effectiveness of our approach with limited data, as the publicly released Cambrian-1 models were trained on millions of samples, requiring immense resources.

\noindent
\textbf{Additional Benchmarks.} Although not the target tasks for our method, we report results on classical visual reasoning benchmarks like POPE~\cite{pope}, GQA~\cite{gqa}, MMMU~\cite{yue2023mmmu}, and VizWiz~\cite{vizwiz} in \cref{tab:extra-bench}. \modelname outperforms LLaVA-1.5, demonstrating its superior visual perception ability without any loss in general reasoning abilities.


\noindent
\textbf{Scalability with VPT.} We report results using the 2.5-stage training setup (PT+VPT+IFT) for LLaVA-1.5~\cite{liu2023improvedllava} and \modelname in \cref{tab:abl_vpt}. Our \modelname outperforms the baselines on average across all CV-Bench tasks, highlighting the scalability of our approach with more training data. 


\noindent
\textbf{Results on BLINK~\cite{fu2024blink} benchmark.} We report results on visual perception tasks under the BLINK benchmark (val set) in \cref{tab:blink}. We also report results using the data-matched Cambrian-1 and RADIO-ViT-L-based LLaVA-1.5 models. We find that our VisPer-LM significantly outperforms all other baselines on average, especially on the Spatial Relation task, demonstrating the effectiveness and generalization of our approach.

\subsection{Ablations}


\noindent
\textbf{Layer Sets for Embedding Losses.} The choice of layers in $\mathbb{D}$, $\mathbb{S}$, and $\mathbb{G}$ for the corresponding embedding losses is a crucial design choice with a significant effect on the performance. \changes{Based on the findings about middle layers in \cref{sec:probe}, we ablate on embedding loss positions only in the middle layers of the LLM.} As shown in \cref{tab:abl-layers}, setting $\mathbb{D}$, $\mathbb{S}$, and $\mathbb{G}$ as $\{8,20\}$, $\{10,18\}$, and $\{12,20\}$, respectively, performs the best overall for the 32-layer Llama3-8b.

\noindent
\textbf{Training Stage for Embedding Optimization.} In Table~\ref{tab:abl_stage}, we observe that adding $\mathcal{L}_\text{emb}$ during IFT results in worse performance than doing so only during the PT stage. We attribute this to the interference of vision-centric supervision with task-aligned natural language supervision during IFT. Additionally, the second row of Table~\ref{tab:abl_stage} shows that using learnable tokens in the sequence~\cite{goyal2024think} without $\mathcal{L}_\text{emb}$ slightly improves performance over the baseline but remains inferior to \modelname.

\begin{table}[!t]
  \centering
  \caption{\textbf{Ablation on the nature of special tokens during IFT.} Keeping $\langle t \rangle $ frozen during IFT aids in keeping their task-specific nature intact, resulting in better performance.}
  \vspace{0.2cm}
  \resizebox{0.8\linewidth}{!}{
  \begin{tabular}{c|cccc|c}
\textbf{$\langle t \rangle $ during IFT} & \textbf{CV-Bench}$^{2D}$ & \textbf{CV-Bench}$^{3D}$ & \textbf{MMStar} & \textbf{RWQA} & \textbf{Avg} \\
\toprule

\rowcolor{azure!10}
frozen & \textbf{58.6} & \textbf{64.2} & \textbf{39.5} & \textbf{57.9} & \textbf{55.1} \\

learnable & 56.9 & 56.1 & 39.0 & 57.3 & 52.3 \\

\bottomrule
\end{tabular}
}
  \vspace{-0.2cm}
    \label{tab:freeze_token}
\end{table}

\begin{table}[t!]
  \centering
  \caption{\textbf{Throughput Analysis.} Our \modelname has superior inference throughout compared to that of the multi-encoder baseline and similar throughput to that of the single-encoder baseline with much better performance.}
  \vspace{0.1cm}
  \resizebox{0.7\linewidth}{!}{
  \begin{tabular}{l|ccc}
inference & single-encoder & multi-encoder & \textbf{\modelname}\\
\toprule

throughput (samples/sec) & \bl{9.92} $\pm$ 0.03 & \re{5.32} $\pm$ 0.02 & \bl{9.86} $\pm$ 0.01 \\


\bottomrule
\end{tabular}

}
  \vspace{-0.2cm}
    \label{tab:throughput}
\end{table}

\begin{table}[t!]
  \centering
  \caption{\textbf{Experiments with SigLIP-VIT-SO400M~\cite{siglip} as the base vision encoder.} Our \modelname improves significantly over LLaVA-1.5~\cite{liu2023improvedllava} on depth and relation reasoning, our target perception tasks.}
  \vspace{0.1cm}
  \resizebox{0.7\linewidth}{!}{
  \begin{tabular}{l|c|ccc}
\textbf{Method} & \textbf{CV-Bench} & Count & Depth & Relation  \\
\toprule

LLaVA-1.5 & 56.3 & 47.8 & 60.3 & 57.4 \\

\rowcolor{azure!10}
\textbf{\modelname} (ours) & 57.3 \textbf{\textcolor{blue}{\scriptsize (+1.0)}} & 48.7 \textbf{\textcolor{blue}{\scriptsize (+0.9)}} & 63.7 \textbf{\textcolor{blue}{\scriptsize (+3.4)}} & 66.3 \textbf{\textcolor{blue}{\scriptsize (+8.9)}} \\

\bottomrule
\end{tabular}

}
  \vspace{-0.2cm}
    \label{tab:siglip}
\end{table}

\noindent
\textbf{Nature of Special tokens during IFT.} As shown in \cref{tab:freeze_token}, keeping the special tokens ($\langle t \rangle $) frozen during IFT performs better as compared to making them learnable. We attribute frozen tokens performing better to the gradients from natural language supervision interfering with the vision-centric information stored in the special tokens during IFT if those are set as learnable parameters.

\noindent
\textbf{Throughput Analysis.} We demonstrate the superior inference throughput of \modelname over the multi-encoder baseline in \cref{tab:throughput}. We record the throughput on a single NVIDIA 80G A100 GPU for a single forward pass on the CV-Bench evaluation set with a batch size of 1. We report the mean and standard deviation across 10 runs. We use the CLIP-ConvNeXT-XXL~\cite{cherti2023reproducible} and Llama3-8b~\cite{dubey2024llama3herdmodels} based models for the throughput analysis.

\noindent
\textbf{Experiments with SigLIP~\cite{siglip}.} In \cref{tab:res}, we achieved performance gains with CLIP-ConvNeXT-XXL, one of the strongest visual encoders, as also found in Cambrian-1~\cite{tong2024cambrian1} (ranked \textbf{second} overall in their Tab. 2). Here, we experiment with SigLIP-ViT-SO400M (ranked \textbf{first} in Cambrian-1~\cite{tong2024cambrian1}) as our base visual encoder. As expected, we notice performance boosts on the perception tasks in CV-Bench in \cref{tab:siglip}, which have a high positive correlation to the depth and seg representation quality (\cref{sec:probe_corr}), underscoring the effectiveness of our approach with various vision backbones.

\section{Conclusion}
\label{sec:conc}

In this work, we probed MLLMs and established a positive correlation between visual representation quality within the LLM and downstream performance. Building on these insights, we introduced \modelname, the first approach to distill knowledge from target encoders into the LLM via predictive embedding optimization during the pre-training stage, complementing the next token prediction objective. We validated that our embedding optimization results in better vision-language alignment before the IFT stage. Through extensive experiments, we demonstrate \modelname's superiority to the corresponding baselines in 
terms of both representation quality and VQA performance. 

\vspace{0.1cm}
\noindent
\textbf{Limitations and Future Work.} With \modelname, we enhanced the quality of representations inside the LLM with guidance from expert visual perception encoders. That said, our work mainly focuses on improving perceptual reasoning without any loss of general reasoning abilities. Incorporating more general-purpose teacher encoders, such as InternViT~\cite{chen2024far, chen2023internvl}, offers a promising pathway to improve general reasoning abilities. Secondly, in this work, we mainly work with the image modality. Applying predictive embedding optimization for low-level information like motion control~\cite{vjepa,2502.02492} while training on videos could improve MLLMs' spatial and temporal reasoning in the future. Lastly, due to resource constraints, we could not scale \modelname to larger LLMs like Llama3-70b~\cite{dubey2024llama3herdmodels} or Qwen2-72b~\cite{qwen2}, and it remains an intriguing experiment.

\vspace{-0.3cm}
\paragraph{Acknowledgements.} We extend our heartfelt gratitude to Fangrui Zhu, Reuben Tan, Min Shi, Bhavika Devnani, Fiona Ryan, and Chieh-Yun Chen for their valuable feedback and insightful discussions. We also sincerely thank the GCR team at Microsoft for their support in resolving frequent infrastructure challenges, which enabled our experimentation. This work was in part supported by NSF CAREER Award \#2239840, and the National AI Institute for
Exceptional Education (Award \#2229873) by National Science Foundation and the Institute of Education Sciences, U.S. Department of Education. Lastly, we thank the ML Center @Georgia Tech and Microsoft Research for supporting this work.

{
    \small
    \bibliographystyle{plain}
    \bibliography{main}
}

\clearpage
\section*{NeurIPS Paper Checklist}

\begin{enumerate}

\item {\bf Claims}
    \item[] Question: Do the main claims made in the abstract and introduction accurately reflect the paper's contributions and scope?
    \item[] Answer: \answerYes{} 
    \item[] Justification: Please refer to \cref{sec:probe} and \cref{sec:exp} for verifying the claims.
    \item[] Guidelines:
    \begin{itemize}
        \item The answer NA means that the abstract and introduction do not include the claims made in the paper.
        \item The abstract and/or introduction should clearly state the claims made, including the contributions made in the paper and important assumptions and limitations. A No or NA answer to this question will not be perceived well by the reviewers. 
        \item The claims made should match theoretical and experimental results, and reflect how much the results can be expected to generalize to other settings. 
        \item It is fine to include aspirational goals as motivation as long as it is clear that these goals are not attained by the paper. 
    \end{itemize}

\item {\bf Limitations}
    \item[] Question: Does the paper discuss the limitations of the work performed by the authors?
    \item[] Answer: \answerYes{} 
    \item[] Justification: Please refer to the end of \cref{sec:conc}.
    \item[] Guidelines:
    \begin{itemize}
        \item The answer NA means that the paper has no limitation while the answer No means that the paper has limitations, but those are not discussed in the paper. 
        \item The authors are encouraged to create a separate "Limitations" section in their paper.
        \item The paper should point out any strong assumptions and how robust the results are to violations of these assumptions (e.g., independence assumptions, noiseless settings, model well-specification, asymptotic approximations only holding locally). The authors should reflect on how these assumptions might be violated in practice and what the implications would be.
        \item The authors should reflect on the scope of the claims made, e.g., if the approach was only tested on a few datasets or with a few runs. In general, empirical results often depend on implicit assumptions, which should be articulated.
        \item The authors should reflect on the factors that influence the performance of the approach. For example, a facial recognition algorithm may perform poorly when image resolution is low or images are taken in low lighting. Or a speech-to-text system might not be used reliably to provide closed captions for online lectures because it fails to handle technical jargon.
        \item The authors should discuss the computational efficiency of the proposed algorithms and how they scale with dataset size.
        \item If applicable, the authors should discuss possible limitations of their approach to address problems of privacy and fairness.
        \item While the authors might fear that complete honesty about limitations might be used by reviewers as grounds for rejection, a worse outcome might be that reviewers discover limitations that aren't acknowledged in the paper. The authors should use their best judgment and recognize that individual actions in favor of transparency play an important role in developing norms that preserve the integrity of the community. Reviewers will be specifically instructed to not penalize honesty concerning limitations.
    \end{itemize}

\item {\bf Theory assumptions and proofs}
    \item[] Question: For each theoretical result, does the paper provide the full set of assumptions and a complete (and correct) proof?
    \item[] Answer: \answerNA{} 
    \item[] Justification: We do not include any theoretical results or claims.
    \item[] Guidelines:
    \begin{itemize}
        \item The answer NA means that the paper does not include theoretical results. 
        \item All the theorems, formulas, and proofs in the paper should be numbered and cross-referenced.
        \item All assumptions should be clearly stated or referenced in the statement of any theorems.
        \item The proofs can either appear in the main paper or the supplemental material, but if they appear in the supplemental material, the authors are encouraged to provide a short proof sketch to provide intuition. 
        \item Inversely, any informal proof provided in the core of the paper should be complemented by formal proofs provided in appendix or supplemental material.
        \item Theorems and Lemmas that the proof relies upon should be properly referenced. 
    \end{itemize}

    \item {\bf Experimental result reproducibility}
    \item[] Question: Does the paper fully disclose all the information needed to reproduce the main experimental results of the paper to the extent that it affects the main claims and/or conclusions of the paper (regardless of whether the code and data are provided or not)?
    \item[] Answer: \answerYes{} 
    \item[] Justification: Please refer to \cref{sec:probe} for details about our probing experiments and to \cref{sec:method} and \cref{subsec:impl} for our main experiments.
    \item[] Guidelines:
    \begin{itemize}
        \item The answer NA means that the paper does not include experiments.
        \item If the paper includes experiments, a No answer to this question will not be perceived well by the reviewers: Making the paper reproducible is important, regardless of whether the code and data are provided or not.
        \item If the contribution is a dataset and/or model, the authors should describe the steps taken to make their results reproducible or verifiable. 
        \item Depending on the contribution, reproducibility can be accomplished in various ways. For example, if the contribution is a novel architecture, describing the architecture fully might suffice, or if the contribution is a specific model and empirical evaluation, it may be necessary to either make it possible for others to replicate the model with the same dataset, or provide access to the model. In general. releasing code and data is often one good way to accomplish this, but reproducibility can also be provided via detailed instructions for how to replicate the results, access to a hosted model (e.g., in the case of a large language model), releasing of a model checkpoint, or other means that are appropriate to the research performed.
        \item While NeurIPS does not require releasing code, the conference does require all submissions to provide some reasonable avenue for reproducibility, which may depend on the nature of the contribution. For example
        \begin{enumerate}
            \item If the contribution is primarily a new algorithm, the paper should make it clear how to reproduce that algorithm.
            \item If the contribution is primarily a new model architecture, the paper should describe the architecture clearly and fully.
            \item If the contribution is a new model (e.g., a large language model), then there should either be a way to access this model for reproducing the results or a way to reproduce the model (e.g., with an open-source dataset or instructions for how to construct the dataset).
            \item We recognize that reproducibility may be tricky in some cases, in which case authors are welcome to describe the particular way they provide for reproducibility. In the case of closed-source models, it may be that access to the model is limited in some way (e.g., to registered users), but it should be possible for other researchers to have some path to reproducing or verifying the results.
        \end{enumerate}
    \end{itemize}

\item {\bf Open access to data and code}
    \item[] Question: Does the paper provide open access to the data and code, with sufficient instructions to faithfully reproduce the main experimental results, as described in supplemental material?
    \item[] Answer: \answerYes{} for data; \answerNo{} for code. 
    \item[] Justification: We use publicly available datasets which can be easily downloaded from the web. Our codebase is built on top of the LLaVA-1.5~\cite{liu2023improvedllava} codebase, and we will release our code with the camera-ready version.
    \item[] Guidelines:
    \begin{itemize}
        \item The answer NA means that paper does not include experiments requiring code.
        \item Please see the NeurIPS code and data submission guidelines (\url{https://nips.cc/public/guides/CodeSubmissionPolicy}) for more details.
        \item While we encourage the release of code and data, we understand that this might not be possible, so “No” is an acceptable answer. Papers cannot be rejected simply for not including code, unless this is central to the contribution (e.g., for a new open-source benchmark).
        \item The instructions should contain the exact command and environment needed to run to reproduce the results. See the NeurIPS code and data submission guidelines (\url{https://nips.cc/public/guides/CodeSubmissionPolicy}) for more details.
        \item The authors should provide instructions on data access and preparation, including how to access the raw data, preprocessed data, intermediate data, and generated data, etc.
        \item The authors should provide scripts to reproduce all experimental results for the new proposed method and baselines. If only a subset of experiments are reproducible, they should state which ones are omitted from the script and why.
        \item At submission time, to preserve anonymity, the authors should release anonymized versions (if applicable).
        \item Providing as much information as possible in supplemental material (appended to the paper) is recommended, but including URLs to data and code is permitted.
    \end{itemize}

\item {\bf Experimental setting/details}
    \item[] Question: Does the paper specify all the training and test details (e.g., data splits, hyperparameters, how they were chosen, type of optimizer, etc.) necessary to understand the results?
    \item[] Answer: \answerYes{} 
    \item[] Justification: Please refer to \cref{subsec:impl}.
    \item[] Guidelines:
    \begin{itemize}
        \item The answer NA means that the paper does not include experiments.
        \item The experimental setting should be presented in the core of the paper to a level of detail that is necessary to appreciate the results and make sense of them.
        \item The full details can be provided either with the code, in appendix, or as supplemental material.
    \end{itemize}

\item {\bf Experiment statistical significance}
    \item[] Question: Does the paper report error bars suitably and correctly defined or other appropriate information about the statistical significance of the experiments?
    \item[] Answer: \answerNo{} 
    \item[] Justification: We conduct all our experiments with a fixed seed, resulting in a deterministic nature of our results.
    \item[] Guidelines:
    \begin{itemize}
        \item The answer NA means that the paper does not include experiments.
        \item The authors should answer "Yes" if the results are accompanied by error bars, confidence intervals, or statistical significance tests, at least for the experiments that support the main claims of the paper.
        \item The factors of variability that the error bars are capturing should be clearly stated (for example, train/test split, initialization, random drawing of some parameter, or overall run with given experimental conditions).
        \item The method for calculating the error bars should be explained (closed form formula, call to a library function, bootstrap, etc.)
        \item The assumptions made should be given (e.g., Normally distributed errors).
        \item It should be clear whether the error bar is the standard deviation or the standard error of the mean.
        \item It is OK to report 1-sigma error bars, but one should state it. The authors should preferably report a 2-sigma error bar than state that they have a 96\% CI, if the hypothesis of Normality of errors is not verified.
        \item For asymmetric distributions, the authors should be careful not to show in tables or figures symmetric error bars that would yield results that are out of range (e.g. negative error rates).
        \item If error bars are reported in tables or plots, The authors should explain in the text how they were calculated and reference the corresponding figures or tables in the text.
    \end{itemize}

\item {\bf Experiments compute resources}
    \item[] Question: For each experiment, does the paper provide sufficient information on the computer resources (type of compute workers, memory, time of execution) needed to reproduce the experiments?
    \item[] Answer: \answerYes{} 
    \item[] Justification: Please refer to \cref{subsec:impl}.
    \item[] Guidelines:
    \begin{itemize}
        \item The answer NA means that the paper does not include experiments.
        \item The paper should indicate the type of compute workers CPU or GPU, internal cluster, or cloud provider, including relevant memory and storage.
        \item The paper should provide the amount of compute required for each of the individual experimental runs as well as estimate the total compute. 
        \item The paper should disclose whether the full research project required more compute than the experiments reported in the paper (e.g., preliminary or failed experiments that didn't make it into the paper). 
    \end{itemize}
    
\item {\bf Code of ethics}
    \item[] Question: Does the research conducted in the paper conform, in every respect, with the NeurIPS Code of Ethics \url{https://neurips.cc/public/EthicsGuidelines}?
    \item[] Answer: \answerYes{} 
    \item[] Justification: To the best of our knowledge, we follow the code of ethics.
    \item[] Guidelines:
    \begin{itemize}
        \item The answer NA means that the authors have not reviewed the NeurIPS Code of Ethics.
        \item If the authors answer No, they should explain the special circumstances that require a deviation from the Code of Ethics.
        \item The authors should make sure to preserve anonymity (e.g., if there is a special consideration due to laws or regulations in their jurisdiction).
    \end{itemize}

\item {\bf Broader impacts}
    \item[] Question: Does the paper discuss both potential positive societal impacts and negative societal impacts of the work performed?
    \item[] Answer: \answerNA{} 
    \item[] Justification: We do not believe our work has any societal impacts except the usual hallucination concern involved with the usage of MLLMs. Our embedding distillation approach does not affect this in any form or way.
    \item[] Guidelines:
    \begin{itemize}
        \item The answer NA means that there is no societal impact of the work performed.
        \item If the authors answer NA or No, they should explain why their work has no societal impact or why the paper does not address societal impact.
        \item Examples of negative societal impacts include potential malicious or unintended uses (e.g., disinformation, generating fake profiles, surveillance), fairness considerations (e.g., deployment of technologies that could make decisions that unfairly impact specific groups), privacy considerations, and security considerations.
        \item The conference expects that many papers will be foundational research and not tied to particular applications, let alone deployments. However, if there is a direct path to any negative applications, the authors should point it out. For example, it is legitimate to point out that an improvement in the quality of generative models could be used to generate deepfakes for disinformation. On the other hand, it is not needed to point out that a generic algorithm for optimizing neural networks could enable people to train models that generate Deepfakes faster.
        \item The authors should consider possible harms that could arise when the technology is being used as intended and functioning correctly, harms that could arise when the technology is being used as intended but gives incorrect results, and harms following from (intentional or unintentional) misuse of the technology.
        \item If there are negative societal impacts, the authors could also discuss possible mitigation strategies (e.g., gated release of models, providing defenses in addition to attacks, mechanisms for monitoring misuse, mechanisms to monitor how a system learns from feedback over time, improving the efficiency and accessibility of ML).
    \end{itemize}
    
\item {\bf Safeguards}
    \item[] Question: Does the paper describe safeguards that have been put in place for responsible release of data or models that have a high risk for misuse (e.g., pretrained language models, image generators, or scraped datasets)?
    \item[] Answer: \answerNo{} 
    \item[] Justification: We shall release the code and models under a non-commercial license upon acceptance to ensure only research use.
    \item[] Guidelines:
    \begin{itemize}
        \item The answer NA means that the paper poses no such risks.
        \item Released models that have a high risk for misuse or dual-use should be released with necessary safeguards to allow for controlled use of the model, for example by requiring that users adhere to usage guidelines or restrictions to access the model or implementing safety filters. 
        \item Datasets that have been scraped from the Internet could pose safety risks. The authors should describe how they avoided releasing unsafe images.
        \item We recognize that providing effective safeguards is challenging, and many papers do not require this, but we encourage authors to take this into account and make a best faith effort.
    \end{itemize}

\item {\bf Licenses for existing assets}
    \item[] Question: Are the creators or original owners of assets (e.g., code, data, models), used in the paper, properly credited and are the license and terms of use explicitly mentioned and properly respected?
    \item[] Answer: \answerYes{} 
    \item[] Justification: We have credited all prior works with citations and will include the corresponding license in our code release upon acceptance.
    \item[] Guidelines:
    \begin{itemize}
        \item The answer NA means that the paper does not use existing assets.
        \item The authors should cite the original paper that produced the code package or dataset.
        \item The authors should state which version of the asset is used and, if possible, include a URL.
        \item The name of the license (e.g., CC-BY 4.0) should be included for each asset.
        \item For scraped data from a particular source (e.g., website), the copyright and terms of service of that source should be provided.
        \item If assets are released, the license, copyright information, and terms of use in the package should be provided. For popular datasets, \url{paperswithcode.com/datasets} has curated licenses for some datasets. Their licensing guide can help determine the license of a dataset.
        \item For existing datasets that are re-packaged, both the original license and the license of the derived asset (if it has changed) should be provided.
        \item If this information is not available online, the authors are encouraged to reach out to the asset's creators.
    \end{itemize}

\item {\bf New assets}
    \item[] Question: Are new assets introduced in the paper well documented and is the documentation provided alongside the assets?
    \item[] Answer: \answerNA{} 
    \item[] Justification: We do not release any new asset.
    \item[] Guidelines:
    \begin{itemize}
        \item The answer NA means that the paper does not release new assets.
        \item Researchers should communicate the details of the dataset/code/model as part of their submissions via structured templates. This includes details about training, license, limitations, etc. 
        \item The paper should discuss whether and how consent was obtained from people whose asset is used.
        \item At submission time, remember to anonymize your assets (if applicable). You can either create an anonymized URL or include an anonymized zip file.
    \end{itemize}

\item {\bf Crowdsourcing and research with human subjects}
    \item[] Question: For crowdsourcing experiments and research with human subjects, does the paper include the full text of instructions given to participants and screenshots, if applicable, as well as details about compensation (if any)? 
    \item[] Answer: \answerNA{} 
    \item[] Justification: The paper does not involve crowdsourcing nor research with human subjects.
    \item[] Guidelines:
    \begin{itemize}
        \item The answer NA means that the paper does not involve crowdsourcing nor research with human subjects.
        \item Including this information in the supplemental material is fine, but if the main contribution of the paper involves human subjects, then as much detail as possible should be included in the main paper. 
        \item According to the NeurIPS Code of Ethics, workers involved in data collection, curation, or other labor should be paid at least the minimum wage in the country of the data collector. 
    \end{itemize}

\item {\bf Institutional review board (IRB) approvals or equivalent for research with human subjects}
    \item[] Question: Does the paper describe potential risks incurred by study participants, whether such risks were disclosed to the subjects, and whether Institutional Review Board (IRB) approvals (or an equivalent approval/review based on the requirements of your country or institution) were obtained?
    \item[] Answer: \answerNA{} 
    \item[] Justification: The paper does not involve crowdsourcing nor research with human subjects.
    \item[] Guidelines:
    \begin{itemize}
        \item The answer NA means that the paper does not involve crowdsourcing nor research with human subjects.
        \item Depending on the country in which research is conducted, IRB approval (or equivalent) may be required for any human subjects research. If you obtained IRB approval, you should clearly state this in the paper. 
        \item We recognize that the procedures for this may vary significantly between institutions and locations, and we expect authors to adhere to the NeurIPS Code of Ethics and the guidelines for their institution. 
        \item For initial submissions, do not include any information that would break anonymity (if applicable), such as the institution conducting the review.
    \end{itemize}

\item {\bf Declaration of LLM usage}
    \item[] Question: Does the paper describe the usage of LLMs if it is an important, original, or non-standard component of the core methods in this research? Note that if the LLM is used only for writing, editing, or formatting purposes and does not impact the core methodology, scientific rigorousness, or originality of the research, declaration is not required.
    \item[] Answer: \answerNA{} 
    \item[] Justification: We did not use LLMs except for training LLM-based MLLMs.
    \item[] Guidelines:
    \begin{itemize}
        \item The answer NA means that the core method development in this research does not involve LLMs as any important, original, or non-standard components.
        \item Please refer to our LLM policy (\url{https://neurips.cc/Conferences/2025/LLM}) for what should or should not be described.
    \end{itemize}

\end{enumerate}

\clearpage
\appendix
\begin{center}{\bf \Large Appendix}\end{center}
\renewcommand{\thetable}{\Roman{table}}
\renewcommand{\thefigure}{\Roman{figure}}
\setcounter{table}{0}
\setcounter{figure}{0}

\Crefname{appendix}{Appendix}{Appendixes}

\noindent
In this appendix, we first share additional ablations, including the effect of the order of different special tokens (\re{$\langle g \rangle$},\bl{$\langle d \rangle$},\epp{$\langle s \rangle$}) in the input sequence to the LLM and the different key input possibilities to the embedding predictor in \cref{sec:addn_abl}. We use CLIP-ViT-L~\cite{clip} and Llama3-8b~\cite{dubey2024llama3herdmodels} as the base vision encoder and decoder LLM, respectively, for the ablations, unless mentioned otherwise.
Next, we demonstrate the effectiveness of our probing setup for encoder-free MLLMs in \cref{sec:probe_cham} and establish a clear correlation between probing performance and CV-Bench accuracy in \cref{sec:probe_corr}. Lastly, we provide qualitative and quantitative analysis on the downstream probing tasks in \cref{sec:probe_down}.

\section{Additional Ablations}
\label{sec:addn_abl}

\begin{table}[t!]
  \centering
  \caption{\textbf{Using additional data during PT v/s embedding optimization.} Our \modelname demonstrates superior performance than the LLaVA-1.5 model trained on with additional ALLaVA-Caption~\cite{chen2024allava} data during the PT stage, underscoring the effectiveness of our approach with limited data. We use CLIP-ConvNeXT-XXL~\cite{cherti2023reproducible,convnext} as the base visual encoder.}
  \vspace{0.1cm}
  \resizebox{1.0\linewidth}{!}{
  \begin{tabular}{c|cc|cccc|c}
\textbf{Method} & \textbf{PT} & \textbf{IFT} & \textbf{CV-Bench}$^{2D}$ & \textbf{CV-Bench}$^{3D}$ & \textbf{MMStar} & \textbf{OK-VQA} & \textbf{Avg} \\

\toprule

LLaVA-1.5 & LLaVA-558K & LLaVA-665k & 60.0 & 56.3 & 37.4 & 56.0 & 52.4 \\

LLaVA-1.5 & LLaVA-558K + ALLaVA-Caption-663K & LLaVA-665k & 56.8 & 60.8 & 37.1 & 57.5 & 53.1 \\

\midrule
\rowcolor{azure!10}
\textbf{\modelname} & LLaVA-558K & LLaVA-665k & \textbf{60.8} & \textbf{62.2} & \textbf{38.5} & \textbf{59.0} & \textbf{55.1} \\

\bottomrule
\end{tabular}

}
  \vspace{-0.4cm}
    \label{tab:extra_pt}
\end{table}

\begin{table}[t!]
  \centering
  \caption{Setting $\mathbf{N}=8$ is optimal with $\mathbf{N}=0$ setting also giving boosts on CV-Bench and MMStar.}
  \vspace{0.1cm}
  \resizebox{0.7\linewidth}{!}{
  \begin{tabular}{c|cccc|c}
$\mathbf{N}$ & \textbf{CV-Bench}$^{2D}$ & \textbf{CV-Bench}$^{3D}$ & \textbf{MMStar} & \textbf{RWQA} & \textbf{Avg} \\
\toprule

single-encoder & 56.0 & 61.0 & 38.8 & 57.8 & 53.4 \\
\midrule

0 & 56.1 & 62.0 & \textbf{40.1} & 56.3 & 53.6 \\


\rowcolor{azure!10}
8 & \textbf{58.6} & \textbf{64.2} & 39.5 & \textbf{57.9} & \textbf{55.1} \\


16 & 56.6 & 63.6 & 37.1 & 54.5 & 52.9 \\

24 & 55.7 & 60.0 & 39.3 & 57.4 & 53.1 \\

\bottomrule
\end{tabular}
}
  \vspace{-0.2cm}
    \label{tab:abl_tokens}
\end{table}

\begin{table}[t!]
  \centering
  \caption{\textbf{Key input to the Embedding Predictor.}. Feeding the tokens corresponding to the system prompt, image, corresponding special tokens, and the text query is optimal.}
  \vspace{0.1cm}
  \resizebox{0.8\linewidth}{!}{
  \begin{tabular}{l|cccc|c}
\textbf{key input to emb. predictor} & \textbf{CV-Bench}$^{2D}$ & \textbf{CV-Bench}$^{3D}$ & \textbf{MMStar} & \textbf{RWQA} & \textbf{Avg} \\
\toprule

{$\langle img \rangle $} $|$ {$\langle t \rangle $} & 53.0 & 54.6 & 38.4 & 56.7 & 50.7 \\

{$\langle sys \rangle $} $|$ {$\langle img \rangle $} $|$ {$\langle t \rangle $} & \textbf{58.7} & 63.0 & 38.8 & 57.4 & 54.5 \\

\rowcolor{azure!10}
{$\langle sys \rangle $} $|$ {$\langle img \rangle $} $|$ {$\langle t \rangle $} $|$ {$\langle txt \rangle $} & 58.6 & \textbf{64.2} & \textbf{39.5} & \textbf{57.9} & \textbf{55.1} \\

\bottomrule
\end{tabular}
}
  \vspace{-0.2cm}
    \label{tab:resampler_inputs}
\end{table}

\noindent
\textbf{Number of Special Tokens.} We analyze the effect of the number of special tokens per target feature ($\langle t \rangle $) on the performance in \cref{tab:abl_tokens}. Setting $\mathbf{N}$ as eight results in the best average performance across benchmarks. Furthermore, \modelname without $\langle t \rangle $ also outperforms the baseline on CV-Bench$^{3D}$ and MMStar, respectively, demonstrating the effectiveness of our embedding optimization.

\noindent
\textbf{Input tokens to Embedding Predictor.} As shown in \cref{tab:resampler_inputs}, we find that including the tokens corresponding to the system prompt in the key input to the embedding predictor is critical for performance. We attribute it to system tokens having high attention scores and effect on the generation~\cite{xiao2023streamingllm}. Therefore, distilling target information into the system tokens is crucial for performance. Moreover, including the text query tokens in the key input to the embedding predictors also results in a slight performance boost as the text tokens hold global image information~\cite{kaduri2024_vision_of_vlms}.

\begin{table}[!t]
  \centering
  \caption{\textbf{Embedding Optimization Modes.} Using the depth, seg, and gen embedding losses simultaneously is optimal.}
  \vspace{0.1cm}
  \resizebox{0.7\linewidth}{!}{
  \begin{tabular}{l|ccc|c}
\textbf{mode} & \textbf{CV-Bench}$^{2D}$ & \textbf{CV-Bench}$^{3D}$ & \textbf{MMStar} & \textbf{Avg} \\
\toprule

LLaVA-1.5 & 56.0 & 61.0 & 38.8 & 51.9 \\
\midrule

\bl{depth} & \textbf{58.6} & 63.5 & 38.8 & 53.6 \\

\epp{seg} & 56.2 & 57.6 & 38.2 & 50.7 \\

\re{gen} & 56.2 & \textbf{65.8} & 39.3 & 53.8 \\

\midrule

\bl{depth} + \epp{seg} & 58.6 & 61.8 & 38.6 & 53.0 \\

\bl{depth} + \re{gen} & 53.6 & 61.8 & 38.8 & 51.4 \\

\epp{seg} + \re{gen} & 54.2 & 60.2 & 39.3 & 51.2 \\

\midrule
\rowcolor{azure!10}
\bl{depth} + \epp{seg} + \re{gen} & \textbf{58.6} & 64.2 & \textbf{39.5} & \textbf{54.1} \\ 

\bottomrule
\end{tabular}
}
  \vspace{-0.2cm}
    \label{tab:abl_mode}
\end{table}

\begin{table}[t!]
  \centering
  \caption{\textbf{Order of different special tokens in the input sequence to the LLM.} Appending the gen, depth, and seg tokens (in that order) in the LLM's input sequence after the image tokens is the optimal setup.}
  \vspace{0.1cm}
  \resizebox{0.8\linewidth}{!}{
  \begin{tabular}{c|cccc|c}

\textbf{$\langle t \rangle $ order} & \textbf{Count}$^\text{2D}$ & \textbf{Depth}$^\text{3D}$ & \textbf{Relation}$^\text{2D}$ & \textbf{Distance}$^\text{3D}$ & \textbf{Overall} \\
\toprule

LLaVA-1.5 & 50.4 & 73.3 & 64.9 & 48.7 & 58.5 \\

\midrule

\bl{$\langle d \rangle $} $|$ \epp{$\langle s \rangle $} $|$ \re{$\langle g \rangle $} & 49.4 & 68.7 & 69.2 & \textbf{56.2} & 59.9 \\

\bl{$\langle d \rangle $} $|$ \re{$\langle g \rangle $} $|$ \epp{$\langle s \rangle $} & \textbf{51.6} & 72.8 & 70.3 & 54.5 & \textbf{61.4} \\

\midrule

\epp{$\langle s \rangle $} $|$ \bl{$\langle d \rangle $} $|$ \re{$\langle g \rangle $} & 48.7 & 71.3 & 65.2 & 52.5 & 58.5 \\

\epp{$\langle s \rangle $} $|$ \re{$\langle g \rangle $} $|$ \bl{$\langle d \rangle $} & 46.7 & 71.3 & \textbf{71.2} & 50.8 & 58.9 \\

\midrule

\rowcolor{azure!10}
\re{$\langle g \rangle $} $|$ \bl{$\langle d \rangle $} $|$ \epp{$\langle s \rangle $} & 51.3 & \textbf{74.2} & 69.4 & 54.3 & \textbf{61.4} \\

\re{$\langle g \rangle $} $|$ \epp{$\langle s \rangle $} $|$ \bl{$\langle d \rangle $} & 50.9 & 68.8 & 70.0 & 50.5 & 59.2 \\

\bottomrule
\end{tabular}
}
  \vspace{-0.2cm}
    \label{tab:order_tokens}
\end{table}

\begin{table}[t!]
  \centering
  \caption{\textbf{Embedding Loss weights during PT.} Setting each embedding loss' weight to 0.5 is optimal.}
  \vspace{0.1cm}
  \resizebox{0.8\linewidth}{!}{
  \begin{tabular}{ccc|ccc|c}
$\lambda_\text{depth}$ & $\lambda_\text{seg}$ & $\lambda_\text{gen}$ & \textbf{CV-Bench}$^{2D}$ & \textbf{CV-Bench}$^{3D}$ & \textbf{MMStar} & \textbf{Avg} \\

\toprule

\multicolumn{3}{c|}{LLaVA-1.5} & 56.0 & 61.0 & 38.8 & 51.9 \\

\midrule

0.10 & 0.10 & 0.10 & \textbf{60.5} & 61.3 & 38.3 & 53.4 \\

0.25 & 0.25 & 0.25 & 56.3 & 59.4 & 37.1 & 50.9 \\

\rowcolor{azure!10}
0.50 & 0.50 & 0.50 & 58.6 & \textbf{64.2} & \textbf{39.5} & \textbf{54.1} \\

0.75 & 0.75 & 0.75 & 57.9 & 59.4 & 37.6 & 51.6 \\

1.00 & 1.00 & 1.00 & 55.8 & 61.7 & 38.1 & 51.9 \\

\bottomrule
\end{tabular}}
  \vspace{-0.2cm}
    \label{tab:loss_w}
\end{table}

\begin{table}[t!]
  \centering
  \caption{\textbf{Ablations on components of embedding losses.} Using both smooth-L1-loss and contrastive loss to compute the final embedding loss is optimal.}
  \vspace{0.1cm}
  \resizebox{0.7\linewidth}{!}{
  \begin{tabular}{cc|ccc|c}
$\mathcal{L}_{\text{sL1}}$ & $\mathcal{L}_{\text{contrastive}}$ & \textbf{CV-Bench}$^{2D}$ & \textbf{CV-Bench}$^{3D}$ & \textbf{MMStar} & \textbf{Avg} \\

\toprule

\checkmark & & 56.8 & 62.3 & 38.3 & 52.5 \\
\rowcolor{azure!10}
\checkmark & \checkmark & \textbf{58.6} & \textbf{64.2} & \textbf{39.5} & \textbf{54.1} \\

\bottomrule
\end{tabular}}
  \vspace{-0.2cm}
    \label{tab:loss}
\end{table}

\noindent
\textbf{Embedding Optimization Mode.} In this ablation study, we evaluate various combinations of embedding losses applied during pretraining (PT). Our results, summarized in \cref{tab:abl_mode}, reveal that the optimal performance is achieved when all three embedding losses—depth, seg, and gen—are used together. Interestingly, we observe that utilizing only depth or gen embedding losses still leads to notable performance improvements, whereas relying solely on seg embedding loss does not yield significant gains. This suggests that different types of target information contribute uniquely to the distillation process. Investigating how the distillation of one type of target information influences the effectiveness of others presents an intriguing direction for future research.

\noindent
\textbf{Order of Special Tokens.} In \cref{tab:order_tokens}, we ablate the order of different special tokens in the LLM's input sequence. We find that $\{$\re{$\langle g \rangle $} $|$ \bl{$\langle d \rangle $} $|$ \epp{$\langle s \rangle $}$\}$ and $\{$\bl{$\langle d \rangle $} $|$ \re{$\langle g \rangle $} $|$ \epp{$\langle s \rangle $}$\}$ show the best performance on CV-Bench~\cite{tong2024cambrian1}. We choose $\{$\re{$\langle g \rangle $} $|$ \bl{$\langle d \rangle $} $|$ \epp{$\langle s \rangle $}$\}$ as our default order due to its performance being better than the baseline on all sub-tasks in CV-Bench.

\noindent
\textbf{Embedding Loss weights.} In \cref{tab:loss_w}, we ablate on different values of $\lambda_\text{depth}$, $\lambda_\text{seg}$, and $\lambda_\text{gen}$ during for the corresponding embedding losses during the pre-training stage. We find that setting each loss weight to 0.5 is optimal.

\noindent
\textbf{Effect of Contrastive Embedding Loss.} In \cref{tab:loss}, analyze the impact of the contrastive loss component within the embedding loss. Our findings show that incorporating the contrastive loss significantly enhances performance, highlighting its positive influence on the model's effectiveness. We keep the smooth L1 loss as a default component to ensure the embedding predictions maintain the same magnitude as the target features, which is crucial for meaningful visualization.

\begin{table}[t!]
  \centering
  \caption{\textbf{Enhanced Projector post Predictive Embedding Optimization.} We observe that the projector features in \modelname are better than those in LLaVA-1.5 after the PT stage (\texttt{proj\_PT}), indicating a positive effect of the embedding losses during the PT stage on the learned projector. We use CLIP-ConvNeXT-XXL and Llama3-8B as the base visual encoder and decoder LLM, respectively.}
  \vspace{0.1cm}
  \resizebox{\linewidth}{!}{
  \begin{tabular}{l|ll|l|ccc|c}
\textbf{Category} & \textbf{{Model}} & \textbf{Module Output Probed}  &  \textbf{Stage} & \multicolumn{4}{c}{\textbf{Probe cosine-sim Scores}} \\
\toprule
& & & & \bl{depth} & \epp{seg} & \re{gen} & {avg} \\
\midrule

\textcolor{gray}{\texttt{vision\_encoder}} & \multicolumn{2}{c|}{\textcolor{gray}{CLIP-ConvNeXT-XXL}} & --- & \textcolor{gray}{0.545} & \textcolor{gray}{0.512} & \textcolor{gray}{0.719} & \textcolor{gray}{0.592} \\
\midrule

\multirow{2}{*}{\texttt{proj\_PT}} 
& LLaVA-1.5 & Projector & PT & 0.531  & 0.495 & \textbf{0.781} & 0.602 \\

& \cellcolor{azure!10}\textbf{\modelname} (ours) & \cellcolor{azure!10}Projector & \cellcolor{azure!10}PT & \cellcolor{azure!10}\textbf{0.554} & \cellcolor{azure!10}\textbf{0.503} & \cellcolor{azure!10}0.776 & \cellcolor{azure!10}\textbf{0.611} \\





\bottomrule
\end{tabular}
}
  \vspaceundertab
    \label{tab:prob_ref}
\end{table}

\begin{figure*}[t!]
\centering
\includegraphics[width=1.\linewidth]{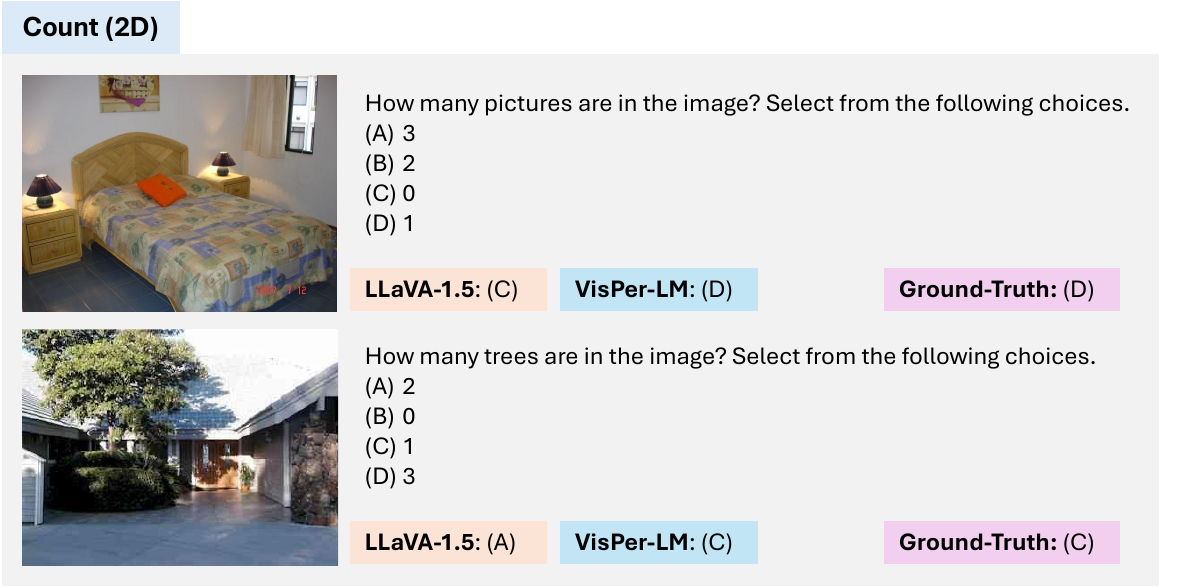} \\
\vspace{-0.1cm}
\caption{\textbf{Qualitative Examples for the Count task in CV-Bench.} Our \modelname can accurately predict the presence of one picture and one tree, unlike LLaVA-1.5~\cite{liu2023improvedllava}.}
\label{fig:count}
\end{figure*}

\begin{figure}[t!]
\centering
\includegraphics[width=1.\linewidth]{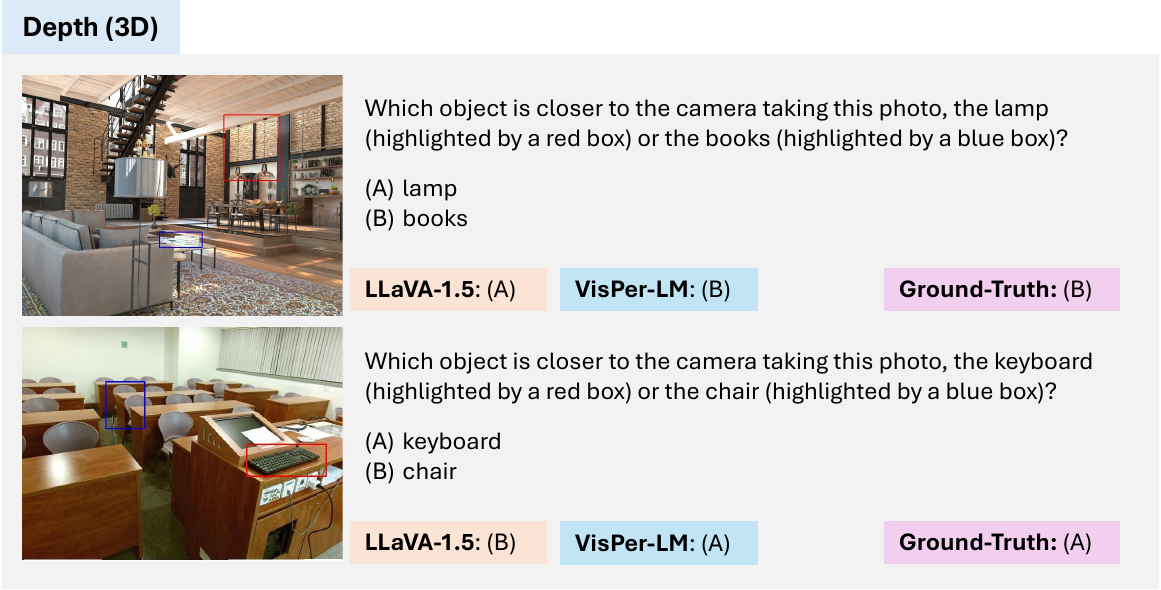} \\
\vspace{0.1cm}
\caption{\textbf{Qualitative Examples for the Depth task in CV-Bench.} Our \modelname can accurately predict that the lamp and keyboard ar closer to the camera in the respective samples.}
\vspace{-0.1cm}
\label{fig:depth}
\end{figure}

\begin{figure}[t!]
\centering
\includegraphics[width=1.\linewidth]{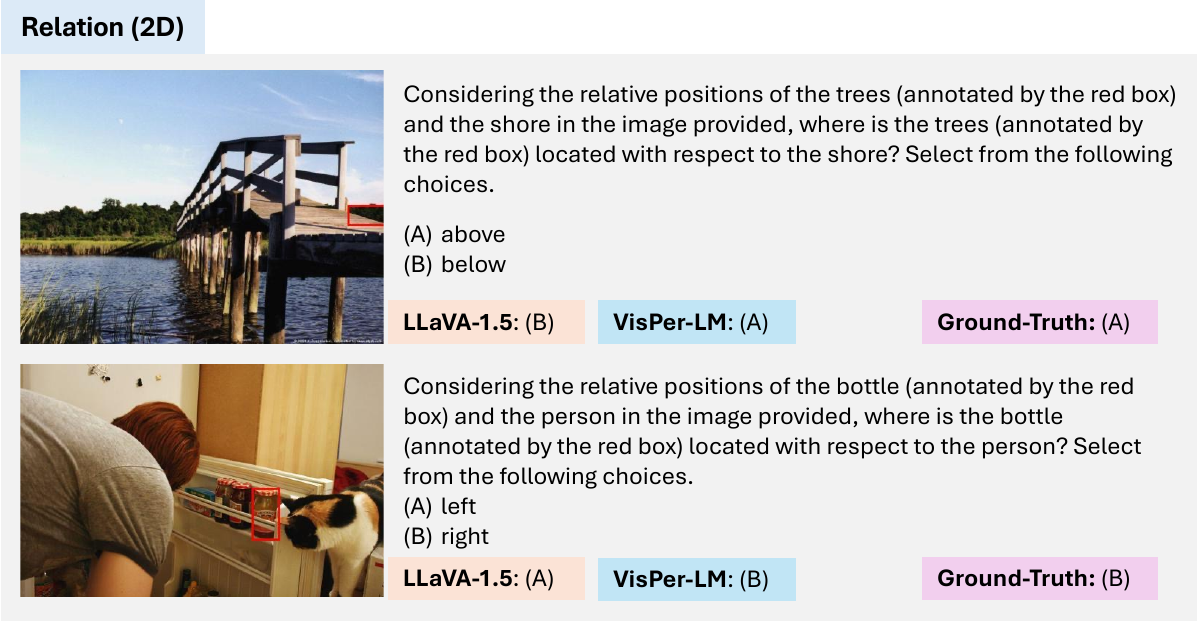} \\
\vspace{0.1cm}
\caption{\textbf{Qualitative Examples for the Relation task in CV-Bench.} Our \modelname can accurately predict that the positions of the trees and the bottle in the respective samples.}
\vspace{-0.2cm}
\label{fig:relation}
\end{figure}

\begin{figure}[t!]
\centering
\includegraphics[width=1.\linewidth]{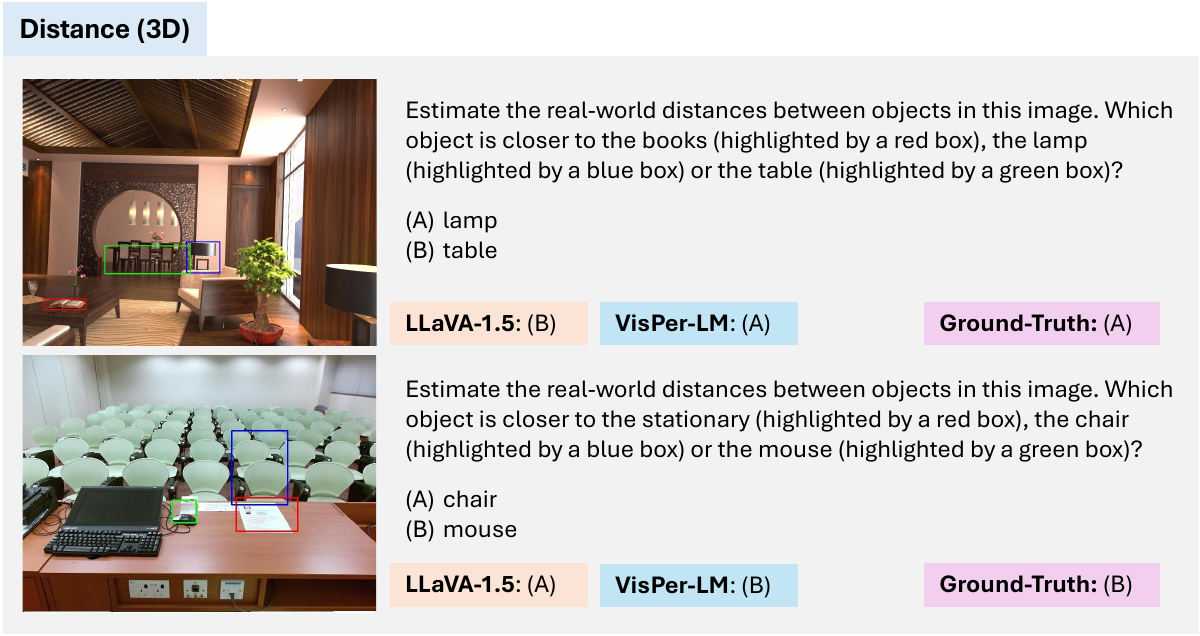} \\
\vspace{0.1cm}
\caption{\textbf{Qualitative Examples for the Distance task in CV-Bench.} Our \modelname can accurately predict that the distances between the respective pair of objects.}
\vspace{-0.2cm}
\label{fig:distance}
\end{figure}

\noindent
\textbf{Qualitative Comparisons.} We provide qualitative comparisons demonstrating the difference between LLaVA-1.5~\cite{liu2023improvedllava} and \textbf{\modelname} for the different tasks in CV-Bench in \cref{fig:count}, \cref{fig:depth}, \cref{fig:relation}, and \cref{fig:distance}.

\begin{figure}[t!]
\centering
\includegraphics[width=0.6\linewidth]{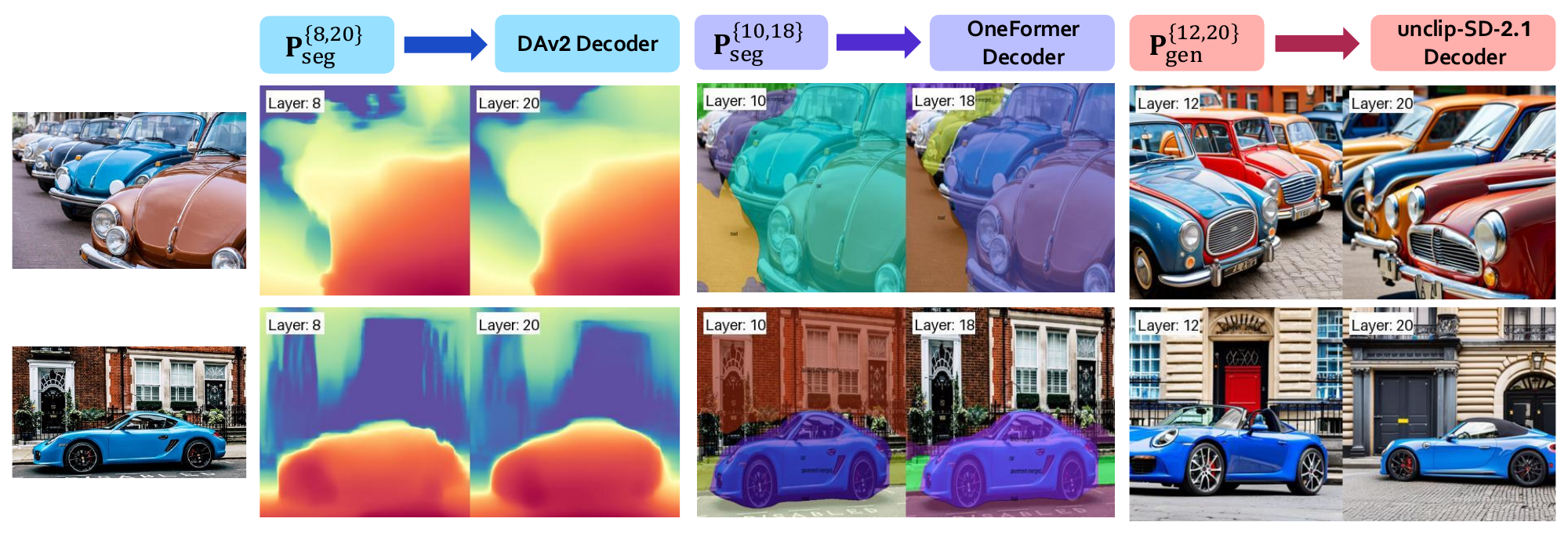} \\
\vspace{0.1cm}
\caption{\textbf{Visualizing Embedding Predictor Outputs after the PT stage.} The quality of the decoded representations indicates the effectiveness of our embedding optimization.}
\vspace{-0.1cm}
\label{fig:aux_vis}
\end{figure}

\noindent
\textbf{Visualizing Embedding Predictions for Target Tasks.} We visualize the visual quality of LLM representations after the PT stage for our \modelname using the decoders from the corresponding target models, as shown in \cref{fig:aux_vis}. We observe that the decoder outputs have good object shape and boundary quality, demonstrating the successful representation optimization with our embedding losses.

\begin{figure}[t!]
\centering
\includegraphics[width=0.6\linewidth]{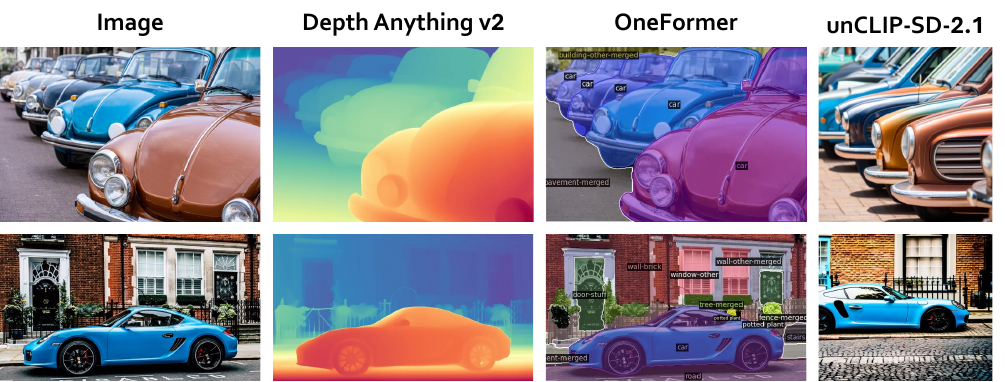} \\
\vspace{0.1cm}
\caption{Ground-truth outputs from the target models used for Probing MLLMs.}
\vspace{-0.1cm}
\label{fig:gt_probe}
\end{figure}

\begin{figure*}[t!]
\centering
\includegraphics[width=1\linewidth]{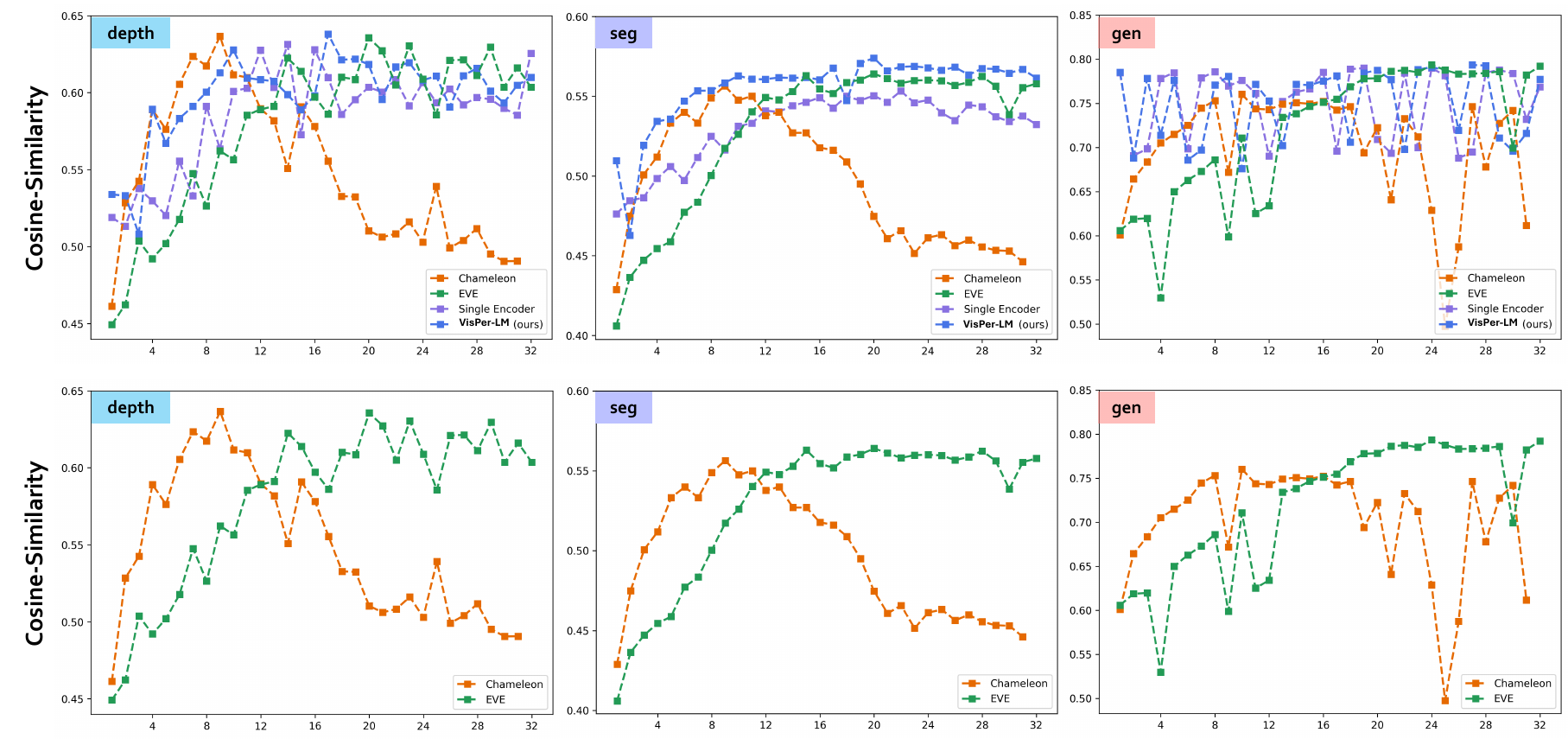} \\
\vspace{-0.1cm}
\caption{\textbf{Probing Encoder-Free MLLMs.} We probe Chameleon-7B~\cite{chameleon} and EVE-7B~\cite{diao2024EVE} models to show the effectiveness of our probing setup for encoder-free MLLMs. In the first row, we compare the mentioned models to LLaVA-1.5 and our \modelname from Sec. \textcolor{cvprblue}{3} of the main text. On the one hand, we notice that EVE-7B shows worse probing performance in the initial layers than LLaVA-1.5, and the representation quality follows an upward trend even in deeper layers, resulting in features of similar quality to LLaVA-1.5, which is also reflected in the CV-Bench accuracy. On the other hand, Chameleon-7B has the best representation quality in the early middle layers, with the quality dropping significantly in the deeper layers. We include only the EVE-7B and Chameleon in the second row for better readability.}
\vspace{-0.3cm}
\label{fig:probe_cham}
\end{figure*}

\begin{table*}[t!]
  \centering
  \caption{\textbf{Evaluating Encoder-Free MLLMs on CV-Bench.} Our \modelname significantly outperforms native encoder free VLMs like EVE-7B~\cite{diao2024EVE} and Chameleon-7B~\cite{chameleon}.}
  \vspace{0.1cm}
  \resizebox{\linewidth}{!}{
  \begin{tabular}{l|ll|cccc|c}
\textbf{Method} & \textbf{Visual Encoder} & \textbf{LLM} & \textbf{Count}$^\text{2D}$ & \textbf{Depth}$^\text{3D}$ & \textbf{Relation}$^\text{2D}$ & \textbf{Distance}$^\text{3D}$ & \textbf{Overall} \\
\toprule

Chameleon~\cite{chameleon} & --- & Llama2-7b & 13.6 & 51.7 & 49.5 & 50.8 & 39.6  \\

EVE~\cite{diao2024EVE} & --- & Vicuna-7b & 52.2 & 61.3 & 69.2 & 53.8 & 58.4  \\

LLaVA-1.5 & CLIP-ConvNeXT-XXL & Llama3-8b & 54.1 & 62.8 & \textbf{69.5} & 49.8 & 58.2 \\

\rowcolor{azure!10}
\textbf{\modelname} (ours) & CLIP-ConvNeXT-XXL & Llama3-8b & \textbf{57.4} & \textbf{71.5} & 66.8 & \textbf{52.8} & \textbf{61.5} \\



\bottomrule
\end{tabular}}
  \vspace{-0.2cm}
    \label{tab:chame_cv}
\end{table*}
\section{Enhanced Projector post Pre-Training}

In this section, we probe the features from the visual encoder and the projector for the LLaVA-1.5 and \modelname models from Sec. \textcolor{cvprblue}{3} of the main text, i.e., using CLIP-ConvNeXT-XXL and the Llama3-8B as the base visual encoder and decoder LLM, respectively. Note that only the projector is trained during the PT stage, while the projector and the LLM are trained during the IFT stage. The visual encoder is kept frozen during both stages.


We compare the probing scores for the feature outputs from the projector module of the LLaVA-1.5 and \modelname models between checkpoints obtained after the PT stage (\texttt{proj\_PT}) of training.  As shown in \cref{tab:prob_ref}, the representations in models under \texttt{proj\_PT} are better for \modelname than LLaVA-1.5, indicating the predictive embedding optimization assisting in learning an improved projector before the IFT stage. 

We observe that the quality of depth and seg representations in LLaVA-1.5 under \texttt{proj\_PT} are worse than those of the base visual encoder (\texttt{vision\_encoder}), indicating a loss of depth/spatial information after the features are fed into the projector in LLaVA-1.5. However, for \modelname, the depth representations have better quality than those from the vision encoder. Note that OLA=VLM's seg representations are slightly worse than those from CLIP-ConvNeXT-XXL but still better than those from LLaVA-1.5.

\section{Probing Encoder-Free MLLMs}
\label{sec:probe_cham}

Following the settings described in Sec. \textcolor{cvprblue}{3} of the main text, we probe all layers from the publicly available encoder-free Chameleon-7B~\cite{chameleon} and EVE-7B~\cite{diao2024EVE} model checkpoints against the depth, seg, and gen target features in \cref{fig:probe_cham}.

We notice that the representation quality for features from EVE-7B is relatively worse in the initial layers, owing to the patch embedding operation that outputs coarse features. However, the probing performance improves in the deeper layers, which could be attributed to the patch-alignment operation~\cite{diao2024EVE}. Notably, EVE-7B uses about 35M~\cite{diao2024EVE} training samples but still lags slightly behind LLaVA-1.5 in performance (\cref{tab:chame_cv}), with our \modelname showing better representation quality and performance, underscoring the effectiveness of our approach.

Unlike EVE-7B, features from Chameleon-7B~\cite{chameleon} have surprisingly low depth and seg representation quality in the deeper layers. Unlike other models, Chameleon-7B follows a steep downward trend in the deeper layers, which we attribute to its relatively low training on VQA data with a heavy focus on text-to-text and text-to-image tasks in the SFT data mixture, which is also evident from its low performance on CV-Bench. Note that during evaluation, we noticed that Chameleon cannot output \textit{option letters} required to evaluate on MCQ-type CV-Bench. Therefore, we leverage Qwen-2.5-72B-Instruct~\cite{qwen2} to extract the option letters corresponding to the sentence output from Chameleon to obtain the numbers for \cref{tab:chame_cv}.

For both the models, gen probes perform relatively better due to the text-aligned nature of the gen target features.

These findings provide critical insights into developing future encoder-free MLLMs, re-establishing the importance of VQA data in the SFT mixture, and supervising the vision tokens inside the LLM from an expert encoder during training. Our proposed embedding optimization opens up a promising avenue to this end.

\section{Correlation between VQA Performance and Probing Performance}
\label{sec:probe_corr}

\begin{tcolorbox}[colback=blue!5!white, colframe=blue!75!black, colframe=black!75!black, title=Takeaways]
\textbf{1.} As shown in \cref{tab:corr}, we observe the highest positive correlation between CV-Bench accuracy and seg/depth probing performance.

\textbf{2.} Moreover, we find a high correlation between the performance on the \textit{Depth/Relation} task and \textit{Count} tasks with the depth and seg probing performance, respectively.

\textbf{3.} As expected, we find low performance correlation with gen probe scores, indicating that gen features act as a good proxy for general visual representation targets.

\textbf{4.} Surprisingly, we observe a negative correlation between the Distance task and all probe scores, which could be attributed to the lack of ``real-world distance" estimation queries in the training dataset.

\end{tcolorbox}

\begin{table*}[t!]
  \centering
  \caption{\textbf{Correlation of CV-Bench performance to probing performance.} We notice the high positive correlation of 0.98 between \bl{depth} probing performance and CV-Bench accuracy. Moreover, we find that performance on the Count and Depth subtasks also highly correlates with the \epp{seg} and \bl{depth} probe cosine-similarity scores, respectively. Additionally, we find a low correlation between the \re{gen} probing performance and CV-Bench performance.}
  \vspace{0.1cm}
  \resizebox{\linewidth}{!}{
  \begin{tabular}{c|cccc|cc|c}

\textbf{Probe Mode} & \textbf{Count}$^\text{2D}$ & \textbf{Depth}$^\text{3D}$ & \textbf{Relation}$^\text{2D}$ & \textbf{Distance}$^\text{3D}$ & \textbf{CV-Bench}$^{2D}$ & \textbf{CV-Bench}$^{3D}$ & \textbf{CV-Bench} \\
\toprule

\bl{depth} & 0.90 & \textbf{0.98} & \textbf{0.70} & -0.31 & \textbf{0.96} & 0.94 & \textbf{0.98} \\

\epp{seg} & \textbf{0.98} & 0.93 & 0.50 & \textbf{-0.15} & 0.91 & \textbf{0.96} & \textbf{0.98} \\

\re{gen} & 0.34 & 0.46 & 0.27 & -0.31 & 0.37 & 0.36 & 0.37 \\ 

\bottomrule
\end{tabular}
}
    \label{tab:corr}
\end{table*}

\begin{figure*}[t!]
\centering
\includegraphics[width=1.\linewidth]{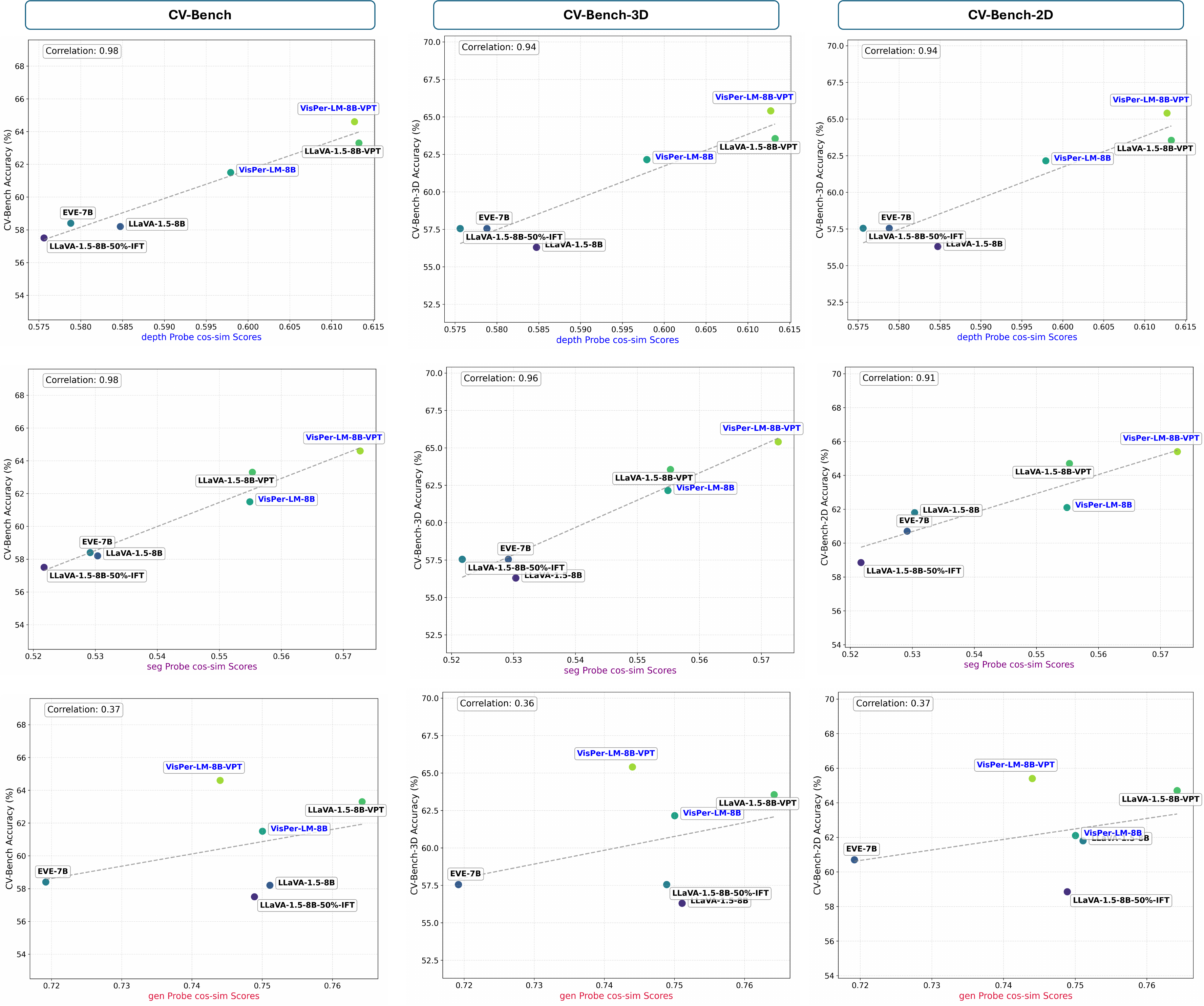} \\
\vspace{-0.1cm}
\caption{\textbf{Plotting Correlation between Probing performance and CV-Bench accuracy.} We find a high correlation between the eval cosine-sim scores and the accuracies on the CV-Bench as well as the 2D and 3D categories under CV-Bench for the \bl{depth} and \epp{seg} probes, validating their choice as the target task to improve the MLLM's perception ability.}
\label{fig:cv_bench_corr}
\end{figure*}

\begin{figure*}[t!]
\centering
\includegraphics[width=1.\linewidth]{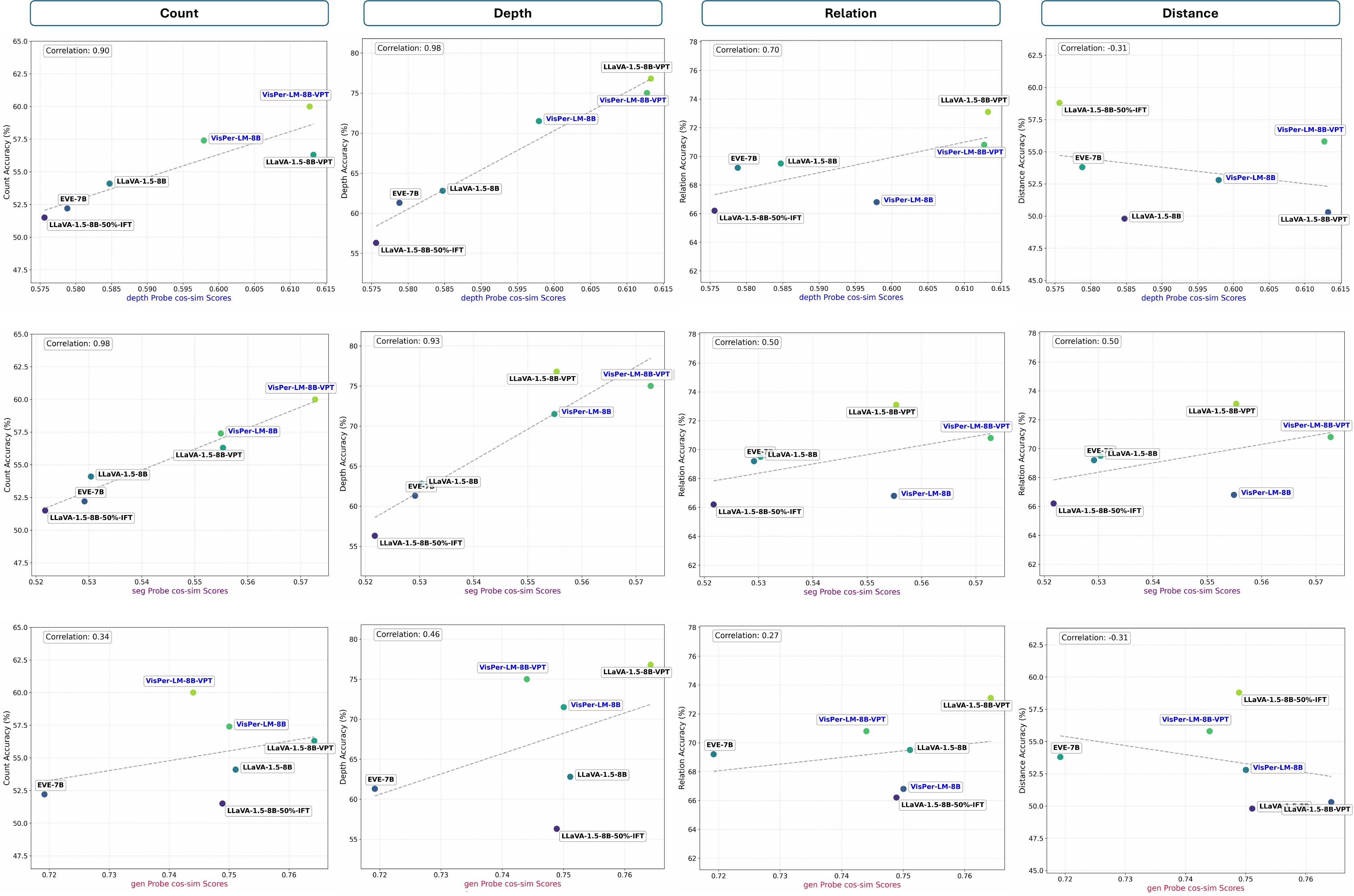} \\
\vspace{-0.1cm}
\caption{\textbf{Plotting Correlation between Probing performance and CV-Bench sub-tasks accuracy.} We find a high correlation between the \bl{depth} eval cosine-sim scores and the Depth/Relation/Count task performances. We make the same observation about \epp{seg} probe scores and the Count/Depth task. Interestingly, we find a low negative correlation between the probe scores and Distance task performance.}
\label{fig:cv_bench_tasks_corr}
\end{figure*}

In this section, we analyze the correlation between the performance on CV-Bench (evaluates models' visual perception ability) and probing performance (measure of model's representation quality), basing our conclusion on trends for six models and their corresponding \bl{depth}/\epp{seg}/\re{gen} probes:

\begin{compactitem}
    \item \textbf{LLaVA-1.5-8B-50\%-IFT}: A LLaVA-1.5 model trained with complete pre-training (PT) but only 50\% completion of the instruction fine-tuning (IFT) stage. It employs CLIP-ConvNeXT-XXL~\cite{cherti2023reproducible,convnext} as the visual encoder and Llama3-8B~\cite{dubey2024llama3herdmodels} as the decoder LLM.
    \item \textbf{LLaVA-1.5-8B}: A LLaVA-1.5 model trained with complete PT+IFT stages. It employs CLIP-ConvNeXT-XXL~\cite{cherti2023reproducible,convnext} as the visual encoder and Llama3-8B~\cite{dubey2024llama3herdmodels} as the decoder LLM.
    \item \bl{\textbf{\modelname-8B}}: Our \modelname model trained with complete PT+IFT stages while using predictive embedding optimization during the PT stage. It employs CLIP-ConvNeXT-XXL~\cite{cherti2023reproducible,convnext} as the visual encoder and Llama3-8B~\cite{dubey2024llama3herdmodels} as the decoder LLM.
    \item \textbf{LLaVA-1.5-8B-VPT}: A LLaVA-1.5 model trained with complete PT+VPT+IFT stages. It employs CLIP-ConvNeXT-XXL~\cite{cherti2023reproducible,convnext} as the visual encoder and Llama3-8B~\cite{dubey2024llama3herdmodels} as the decoder LLM.
    \item \bl{\textbf{\modelname-8B-VPT}}: Our \modelname model trained with complete PT+VPT+IFT stages using predictive embedding optimization during the PT stage. It employs CLIP-ConvNeXT-XXL~\cite{cherti2023reproducible,convnext} as the visual encoder and Llama3-8B~\cite{dubey2024llama3herdmodels} as the decoder LLM.
    \item {\textbf{EVE-7B}}: The publicly available EVE-7B model checkpoint from \cite{diao2024EVE}. It does not use a visual encoder and is built on Vicuna-7B~\cite{vicuna2023}
\end{compactitem}

We follow the same settings as mentioned in Sec. \textcolor{cvprblue}{3} and Sec. \textcolor{cvprblue}{5} of the main text to train our probes and models, respectively.

\textbf{Probing Mode: \bl{depth}.} We observe a high correlation of 0.98 between the CV-Bench accuracy and depth probe eval cosine-sim scores, as shown in \cref{fig:cv_bench_corr}. It is critical proof of our claim that improved visual representations and the corresponding VQA performance go hand-in-hand. We also find a correlation with a 0.98 and 0.80 factor between the accuracy on the Depth and Relation tasks (in CV-bench), respectively, with the depth probe scores in \cref{fig:cv_bench_tasks_corr}, explaining the gains achieved by \modelname. 

\textbf{Probing Mode: \epp{seg}.} We also observe a high correlation of 0.98 between the CV-Bench accuracy and seg probe eval cosine-sim scores, as shown in \cref{fig:cv_bench_corr}. The high correlation indicates that CV-Bench requires strong perception abilities, validating our decision to adopt it as the primary benchmark for our work. We also find a correlation of 0.98 between the accuracy on the Count task (in CV-bench) with the seg probe scores in \cref{fig:cv_bench_tasks_corr}. 

\textbf{Probing Mode: \re{gen}.} We observe that the gen probe scores do not correlate highly with CV-Bench or any of its tasks, as shown in \cref{fig:cv_bench_corr} and \cref{fig:cv_bench_tasks_corr}. We attribute this finding to the generative features' text-aligned nature, which is not optimal for developing the model's perception abilities.

\section{Evaluating Probes on Downstream Tasks}
\label{sec:probe_down}

We evaluate the probes trained against the target features on the corresponding downstream target tasks, i.e., image generation, depth estimation and image segmentation. To obtain the predictions, we feed the outputs from the probes into the decoder from the corresponding target model. Specifically, we report the FID~\cite{Seitzer2020FID} scores on 5k images from COCO-val2017~\cite{coco} for gen probes, accuracy on the DA-2K~\cite{depth_anything_v2} benchmark for depth probes, and mIoU~\cite{jain2023oneformer} on the COCO-val2017~\cite{coco} set for seg probes. We average the scores over all layers for easier comparison. As shown in \cref{tab:probe_task}, probes trained for our \modelname outperform those for the baseline LLaVA-1.5~\cite{liu2023improvedllava} model across all probing tasks, proving the improved visual representation quality owing to our embedding optimization approach.

\begin{table}[t!]
  \centering
  \caption{\textbf{Quantitative Evaluation on target probing task.} Probes trained for our \modelname perform significantly better as compared to the probes trained on baseline LLaVA-1.5~\cite{liu2023improvedllava}.}
  \vspace{0.1cm}
  \resizebox{0.8\linewidth}{!}{
  \begin{tabular}{l|ccc}
\textbf{Probed Model} & \textbf{FID}~\cite{Seitzer2020FID} ($\downarrow$) & \textbf{DA-2K \% Acc.}~\cite{depth_anything_v2} ($\uparrow$) & \textbf{\% mIoU}~\cite{jain2023oneformer} ($\uparrow$) \\
\toprule

LLaVA-1.5 & 23.1 & 66.4 & 39.3 \\

\rowcolor{azure!10}
\textbf{\modelname} (ours)  & \textbf{22.4} & \textbf{77.8} & \textbf{45.4}  \\

\midrule

\textcolor{gray}{Target Encoder} & \textcolor{gray}{18.1} & \textcolor{gray}{97.3} & \textcolor{gray}{64.5} \\
\bottomrule

\end{tabular}}
  \vspace{-0.2cm}
    \label{tab:probe_task}
\end{table}

\begin{figure}[t!]
\centering
\includegraphics[width=1\linewidth]{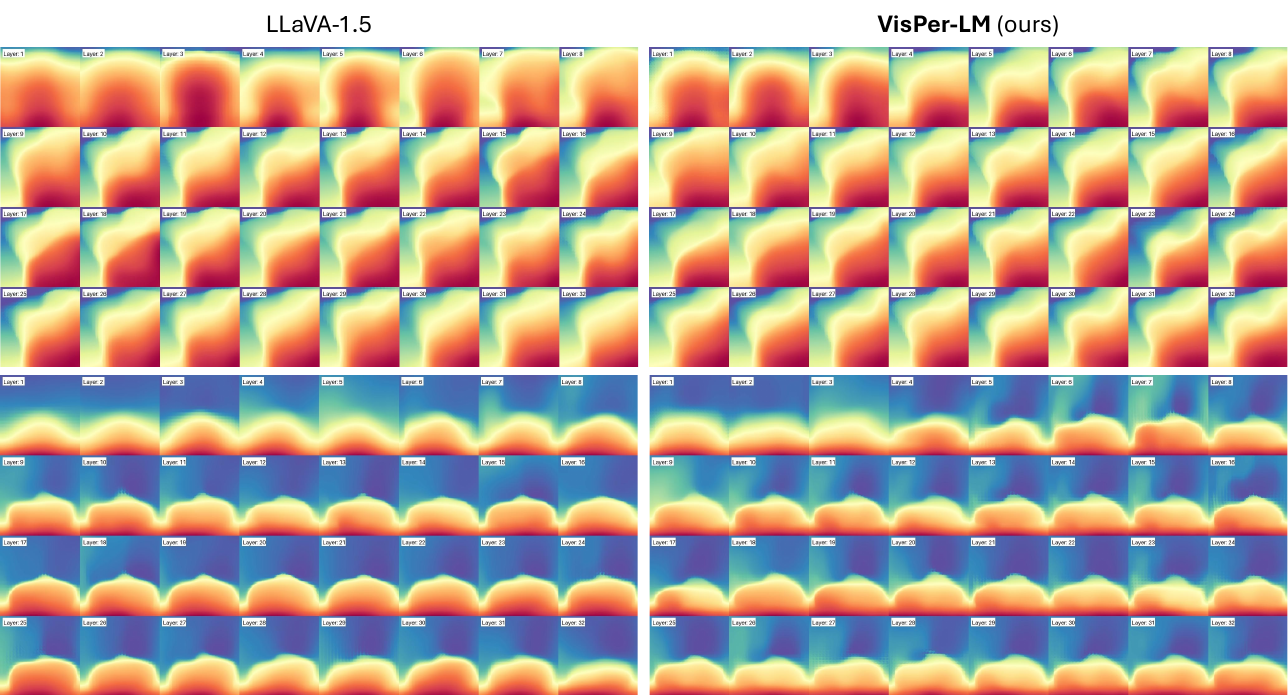} \\
\vspace{0.1cm}
\caption{\textbf{Layerwise visualizations for the \bl{depth} probes.} For LLaVA-1.5~\cite{liu2023improvedllava}, the probes generate blob-like outputs up to the eighth layer, with visualizations progressively improving in the middle layers, aligning with the findings presented in Sec. \textcolor{cvprblue}{3} of the main text. Notably, probes trained on \modelname begin producing distinguishable object shapes and boundaries as early as the third layer, attributed to the incorporation of embedding losses leading to improved representations in the initial layers.}
\vspace{-0.2cm}
\label{fig:d_probe}
\end{figure}

\vspace{0.1cm}
\noindent
\textbf{Visualizing Probe Outputs.} We present visualizations for the outputs for the probes trained on the single-encoder LLaVA-1.5 model in Sec. \textcolor{cvprblue}{3} in the main text. We provide the ground-truth visualizations from the teacher models in \cref{fig:gt_probe}.

\begin{figure}[t!]
\centering
\includegraphics[width=1\linewidth]{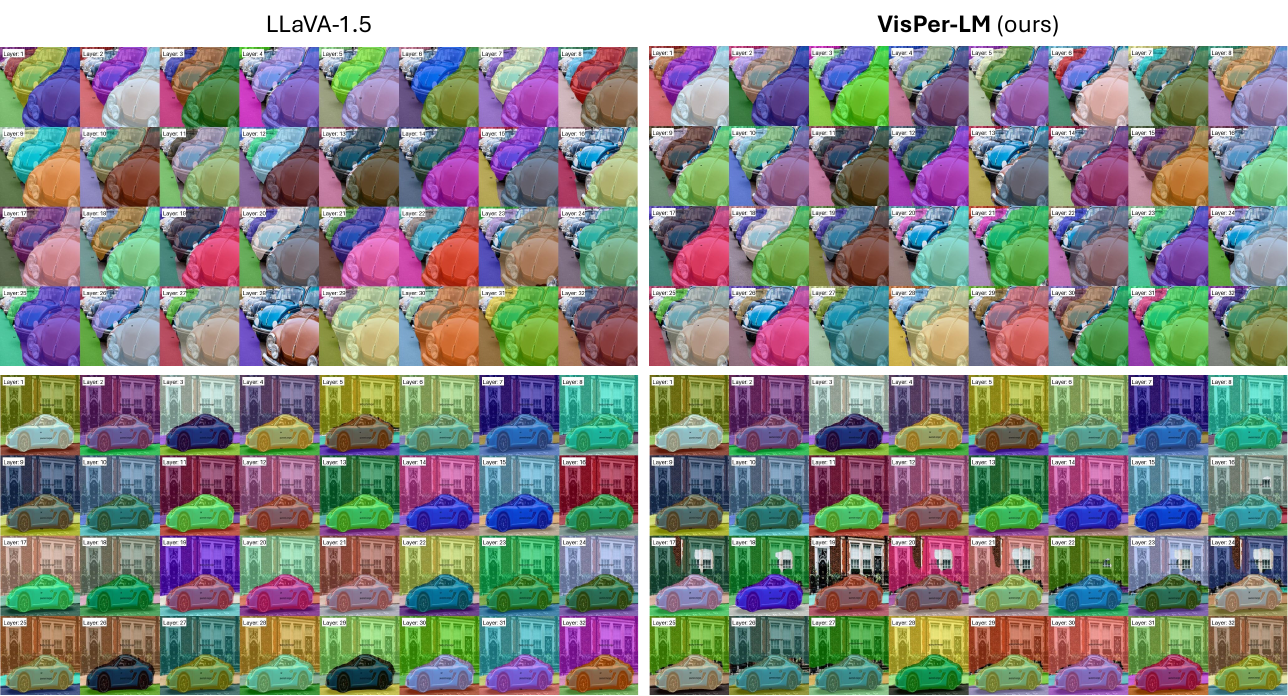} \\
\vspace{0.1cm}
\caption{\textbf{Layerwise visualizations for the \epp{seg} probes.} 
The LLaVA-1.5 probes often fail to segment the third car in the background for the first sample during the initial layers (layers two to eight), whereas the \modelname probes demonstrate relatively better performance in this scenario.}
\label{fig:s_probe}
\end{figure}

\begin{figure}[t!]
\centering
\includegraphics[width=1\linewidth]{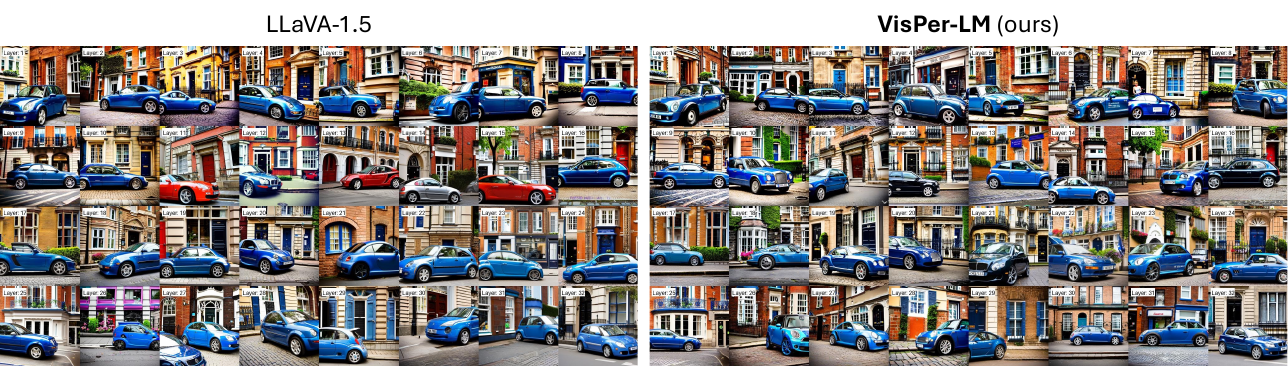} \\
\vspace{0.1cm}
\caption{\textbf{Layerwise visualizations for the \epp{gen} probes.} The probe outputs for both the models are of fairly good quality.}
\vspace{-0.3cm}
\label{fig:g_probe}
\end{figure}

As shown in \cref{fig:d_probe}, the probe visualizations for the first eight layers of LLaVA-1.5 exhibit blob-like patterns, while the later layers progressively enhance the shape and boundary details of the foreground objects. In contrast, the probe visualizations for our \modelname demonstrate improved object shapes and boundaries starting as early as layer-4, consistent with the findings on representation quality and layer-wise trends discussed in Sec. \textcolor{cvprblue}{3} of the main text. Additionally, we present probe visualizations for the seg and gen representations in \cref{fig:s_probe} and \cref{fig:g_probe}, respectively.


\end{document}